\documentclass[lettersize,journal]{IEEEtran}
\usepackage{amsmath,amsfonts}
\usepackage{algorithmic}
\usepackage{algorithm}
\usepackage{array}
\usepackage[caption=false,font=normalsize,labelfont=sf,textfont=sf]{subfig}
\usepackage{textcomp}
\usepackage{stfloats}
\usepackage{url}
\usepackage{verbatim}
\usepackage{graphicx}
\usepackage{cite}
\usepackage{multirow}
\usepackage{adjustbox}
\usepackage{xcolor}

\usepackage{amsmath}
\usepackage{amssymb}
\usepackage{array}
\usepackage{amsthm}

\usepackage{booktabs}

\usepackage{colortbl}
\usepackage[table]{xcolor}
\usepackage{bbding}
\usepackage{relsize}
\usepackage[utf8]{inputenc} % allow utf-8 input
\usepackage[T1]{fontenc}    % use 8-bit T1 fonts
\usepackage{hyperref}       % hyperlinks
\usepackage{nicefrac}       % compact symbols for 1/2, etc.
\usepackage{microtype}      % microtypography
\usepackage{tabularx}
\usepackage{siunitx}       % (可选) 为了更好的对齐小数点，但此处我们用 \pm，故不强制
\newcommand{\result}[2]{{\mathlarger{#1}}{\color{gray}\mathsmaller{\pm #2}}}
\begin{document}

\title{InkDrop: Invisible Backdoor Attacks Against Dataset Condensation}

\author{He Yang, Dongyi Lv, Song Ma, Wei Xi,~\IEEEmembership{Member,~IEEE,} Zhi Wang, Hanlin Gu, Yajie Wang
        % <-this % stops a space
\thanks{This paper was produced by the IEEE Publication Technology Group. They are in Piscataway, NJ.}% <-this % stops a space
\thanks{Manuscript received April 19, 2021; revised August 16, 2021.}}

\markboth{Journal of \LaTeX\ Class Files,~Vol.~14, No.~8, August~2021}%
{Shell \MakeLowercase{\textit{et al.}}: A Sample Article Using IEEEtran.cls for IEEE Journals}


\maketitle

\begin{abstract}
Dataset Condensation (DC) is a data-efficient learning paradigm that synthesizes small yet informative datasets, enabling models to match the performance of full-data training. However, recent work exposes a critical vulnerability of DC to backdoor attacks, where malicious patterns (\textit{e.g.}, triggers) are implanted into the condensation dataset, inducing targeted misclassification on specific inputs. Existing attacks always prioritize attack effectiveness and model utility, overlooking the crucial dimension of stealthiness. To bridge this gap, we propose InkDrop, which enhances the imperceptibility of malicious manipulation without degrading attack effectiveness and model utility. InkDrop leverages the inherent uncertainty near model decision boundaries, where minor input perturbations can induce semantic shifts, to construct a stealthy and effective backdoor attack. Specifically, InkDrop first selects candidate samples near the target decision boundary that exhibit latent semantic affinity to the target class. It then learns instance-dependent perturbations constrained by perceptual and spatial consistency, embedding targeted malicious behavior into the condensed dataset. Extensive experiments across diverse datasets validate the overall effectiveness of InkDrop, demonstrating its ability to integrate adversarial intent into condensed datasets while preserving model utility and minimizing detectability. Our code is available at \url{https://github.com/lvdongyi/InkDrop}.

\end{abstract}


\section{Introduction}
\IEEEPARstart{D}{ataset} Condensation (DC) is an emerging paradigm that aims to synthesize a compact set of representative samples while preserving the training utility of large-scale datasets~\cite{wang2018dataset,zhao2020dataset,he2024multisize,qi2024fetch,zhang2024m3d}. It distills core data characteristics into a reduced form, enabling efficient model training with low memory and computation costs~\cite{sun2024diversity,su2024d,du2024diversity}. DC also enhances privacy by removing the need to share raw data, making it well-suited for collaborative and distributed learning where transferring original datasets is impractical~\cite{dong2022privacy}. Despite its efficiency, DC introduces critical security risks~\cite{sun2024study,zhang2024risk,xue2025towards,zhou2024beard,wu2025dd}. The condensation process can be exploited to implant backdoor triggers into synthetic samples~\cite{liu2023backdoor,chung2024rethinking,zheng2023rdm,arazzi2025secure}. These samples directly influence model behavior, making even minor manipulations sufficient to embed persistent malicious functions. Such vulnerabilities compromise the integrity of downstream models and are especially concerning in safety-critical applications.

\begin{figure}
    \centering
    \includegraphics[width=1\linewidth]{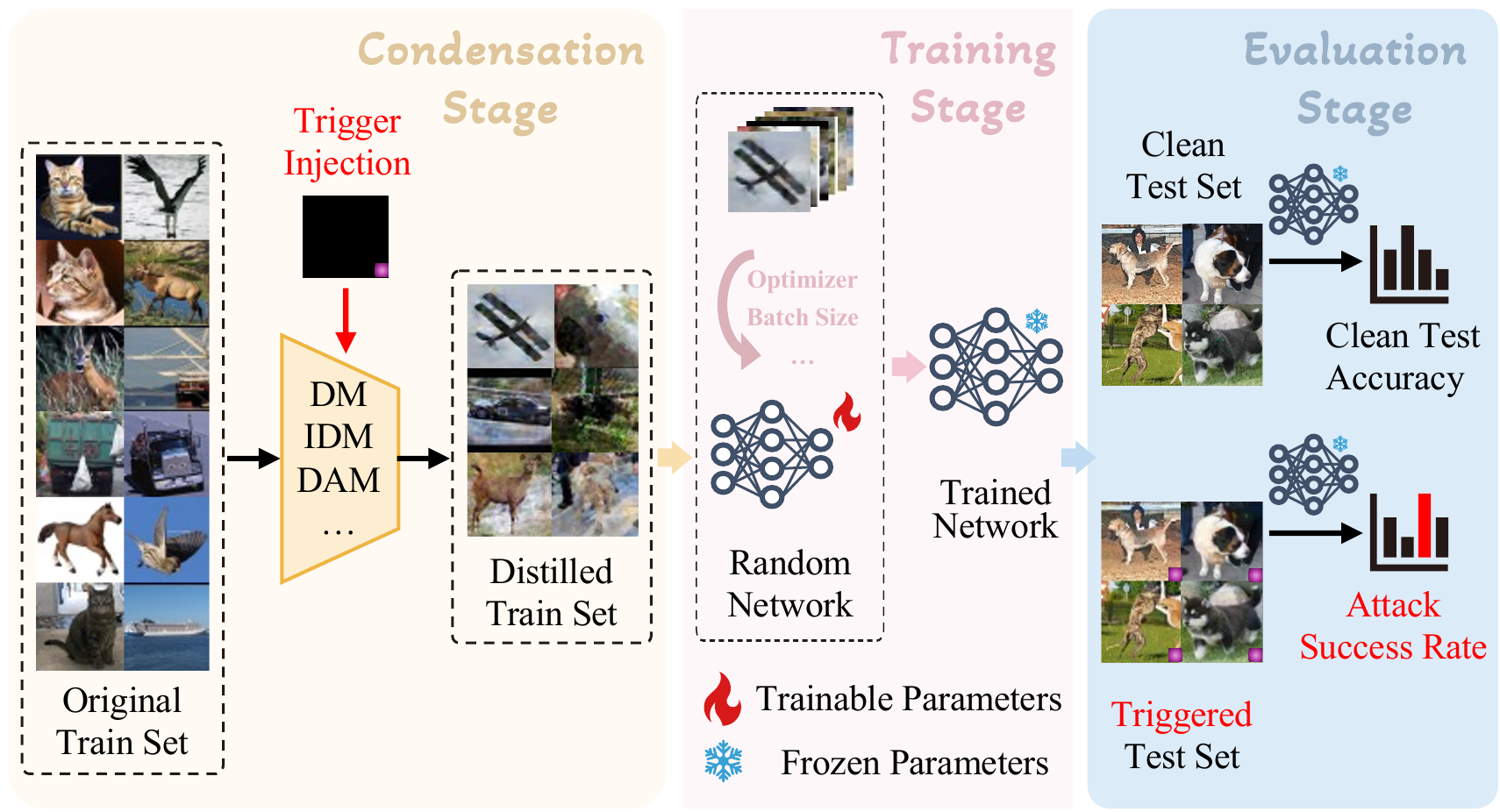}
    \caption{Illustration of backdoor attacks against dataset condensation.}
    \label{fig:backdoordc}
    \vspace{-9pt}
\end{figure}

Recent studies~\cite{liu2023backdoor,chung2024rethinking,zheng2023rdm,yang2025dark,chai2024adaptive} demonstrate that adversarial triggers embedded in synthetic data can persist through downstream training and induce targeted misclassifications at inference (See Figure~\ref{fig:backdoordc}). Among the earliest explorations in this domain is the Naive Attack~\cite{liu2023backdoor}, which demonstrates that simply overlaying a fixed visual pattern onto clean samples before condensation can effectively embed persistent backdoor behavior. Although conceptually simple, Naive Attack provides a foundational proof, validating the feasibility of backdoor injection within the condensation pipeline. Building on this foundation, more advanced strategies have been proposed to amplify the attack effectiveness. A prominent example is Doorping~\cite{liu2023backdoor}, which leverages a bilevel optimization framework to jointly optimize the synthetic dataset and the backdoor trigger. This approach more effectively preserves trigger semantics throughout the condensation process, resulting in substantially higher attack success rates. More recent advances shift toward a theoretical lens, leveraging kernel-based methods to analyze backdoor vulnerabilities in DC. The Simple-Trigger~\cite{chung2024rethinking} minimizes the Generalization Gap, defined as the discrepancy between the expected loss over the full manipulated dataset and the empirical loss on a sampled subset, promoting backdoor generalization beyond seen samples. The Relax-Trigger~\cite{chung2024rethinking}  additionally reduces projection loss, which measures the extent to which a model trained on distilled data generalizes to the original clean distribution, and conflict loss, which captures feature-space interference between clean and poisoned samples. This joint optimization enhances the stability and transferability of the backdoor attack.

\begin{figure}
    \centering
    \includegraphics[width=1\linewidth]{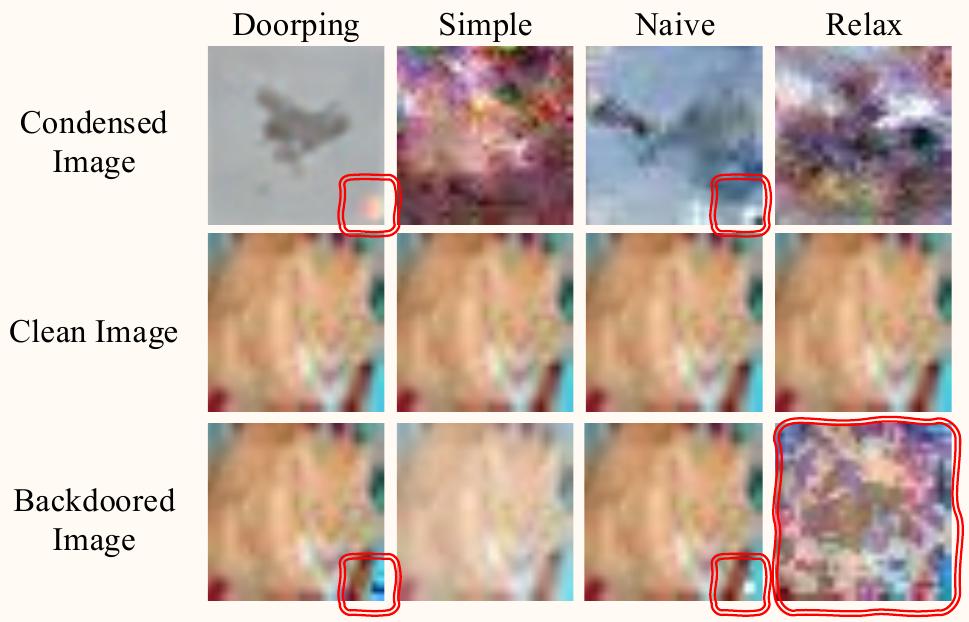}
    \caption{Visualization of the stealth limitations in existing backdoor condensation methods. Both synthetic data and trigger-injected test samples exhibit visible artifacts or unnatural structures, exposing the trigger and undermining the attack's stealthiness.}
    \label{fig:steathiness illustration}
    \vspace{-9pt}
\end{figure}

Existing approaches always prioritize optimizing model behavior and attack efficacy, while overlooking a critical and underexplored limitation: \textit{\textbf{stealthiness}}, either in the synthetic data or in triggered test samples. During the condensation process, synthetic data often contain unnatural patterns that expose the presence of the trigger, compromising the covert nature of the attack (See Figure~\ref{fig:steathiness illustration}). At inference time, the inserted triggers may produce noticeable visual artifacts, making them easily identifiable by human observers or basic anomaly detection systems. This lack of visual subtlety significantly weakens the practical applicability of these attacks. To ensure long-term persistence and real-world viability, future backdoor strategies must account for both functional effectiveness and visual inconspicuity.

Therefore, we propose an invisible backdoor attack InkDrop against DC. InkDrop is motivated by the key observation that models exhibit intrinsic uncertainty near decision boundaries, where samples are highly sensitive to small perturbations and susceptible to label shifts. Leveraging this property, InkDrop initiates its attack by identifying source-class samples that display latent affinity to the target class, as indicated by slightly elevated confidence in the target class dimension. These boundary-adjacent samples form a candidate pool that is inherently vulnerable to misclassification, offering a natural foothold for backdoor injection.

Although these candidate samples are inherently susceptible, embedding triggers that are both effective and stealthy poses a significant challenge. Subtle perturbations risk being attenuated or removed during the condensation process, while stronger triggers often introduce visible artifacts that reveal the attack. Existing approaches often fail to strike a balance between attack success and stealthiness. Consequently, achieving this balance remains a fundamental obstacle for stealthy and reliable backdoor attacks against DC.

To address this challenge, InkDrop introduces a learnable attack model designed to generate customized, low-perceptibility perturbations for each input. The attack model is trained under four complementary objectives: a contrastive InfoNCE loss to align poisoned features with the target class while repelling clean features; an Earth Mover’s Distance loss to align output distributions with target soft labels;  L2 regularization to limit perturbation strength; and an LPIPS perceptual loss to maintain visual consistency. These constraints jointly ensure that the generated triggers are both stealthy and effective. The attack model then applies these instance-dependent triggers to candidate samples, which are combined with clean target-class data and fed into the condensation pipeline, embedding the backdoor covertly into the synthetic dataset.

Main contributions of this paper are summarized below: 

\begin{itemize}
    \item 
    We present InkDrop, a stealthy backdoor attack framework targeting dataset condensation, providing the first comprehensive investigation into jointly optimizing attack effectiveness, model utility, and stealthiness.
    \item 
    InkDrop introduces an instance-dependent attack model that generates visually imperceptible yet effective perturbations via multi-objective optimization, enabling robust and stealthy backdoor injection without compromising condensation quality.
    \item 
    Empirical results on four standard benchmarks show that InkDrop consistently achieves superior trade-offs compared to prior methods, effectively balancing stealthiness, attack effectiveness, and model utility.
\end{itemize}

\section{Related Work}
\subsection{Dataset Condensation}

Dataset condensation (DC)~\cite{lei2023comprehensive,yu2023dataset,liu2023dream} aims to generate a compact synthetic dataset that retains the training utility of the original. Among various paradigms, distribution matching (DM)-based methods are prominent for their scalability, generality, and empirical effectiveness. These methods align feature-level statistics between real and synthetic data, commonly using the maximum mean discrepancy (MMD) to match distributions within a low-dimensional embedding space. A typical objective that minimizes the MMD between augmented feature embeddings of real samples $\mathcal{T}$ and synthetic samples $\mathcal{S}$ is: $\min_{\mathcal{S}} \mathbb{E}_{\boldsymbol{\theta} \sim P_{\boldsymbol{\theta}}} \left\| \frac{1}{|\mathcal{T}|} \sum_{i} \psi(\mathcal{A}(\boldsymbol{x}_i)) - \frac{1}{|\mathcal{S}|} \sum_{j} \psi(\mathcal{A}(\boldsymbol{s}_j)) \right\|^2$, where $\psi$ denotes randomly initialized encoders and $\mathcal{A}(\cdot)$ represents a differentiable augmentation operator. This formulation encourages the synthetic dataset to retain the statistical characteristics of the original data, ensuring generalization across diverse model initializations.

Subsequent extensions, such as IDM~\cite{zhao2023improved} and DAM~\cite{sajedi2023datadam}, enhance class-conditional alignment through kernel-based moment matching, adaptive feature regularization, and encoder updates, yielding improved performance. IDM introduces practical enhancements to the original distribution matching framework, incorporating progressive feature extractor updates, stronger data augmentations, and dynamic class balancing to improve generalization. In parallel, DAM leverages attention map alignment to better preserve spatial semantics, guiding synthetic samples to activate similar regions as real data while maintaining computational efficiency. These methods advance the state of dataset condensation by demonstrating that richer supervision and adaptive training dynamics are critical for generating high-fidelity synthetic datasets.

\subsection{Backdoor against Dataset Condensation}
Backdoor attacks aim to manipulate model behavior at inference time by injecting carefully crafted triggers into a subset of training data. When effective, the model performs normally on clean inputs but consistently misclassifies inputs containing the trigger. While extensively studied in standard supervised learning, backdoor attacks in the context of dataset condensation have only recently received attention. A pioneering study by Liu et al.~\cite{liu2023backdoor} by poisoning real data prior to distillation. Their Naive Attack appends a fixed trigger to target-class samples before condensation, but suffers from trigger degradation and reduced attack efficacy due to the synthesis process. To address this, Doorping employs a bilevel optimization scheme that jointly refines the trigger and the synthetic data. Although more effective, it incurs substantial computational overhead. More recently, Chung et al.~\cite{chung2024rethinking} provide a kernel-theoretic perspective on backdoor persistence in condensation. They propose Simple-Trigger, which minimizes the generalization gap of the backdoor effect, and Relax-Trigger, which further reduces projection and conflict losses for improved robustness. 

However, these methods prioritize either attack effectiveness or model utility, while often neglecting stealthiness. In contrast, we introduce InkDrop, a unified framework that jointly optimizes for attack effectiveness, model utility, and visual stealthiness. % By leveraging instance-dependent trigger generation and stealth-aware integration into existing DC frameworks, InkDrop effectively balances these competing objectives, enabling robust and inconspicuous backdoor injection in condensed datasets.

\section{Methodology}

% To enhance clarity and facilitate understanding of the proposed method, we provide a detailed list of notations used throughout the paper. This summary, presented in Table~\ref{tab:notations}, serves as a concise reference for the key symbols and terms introduced in our framework.

To enhance clarity and facilitate understanding of the proposed method, we provide a detailed list of notations used throughout the paper, presented in Table~\ref{tab:notations}.

\begin{table}[!ht]
\centering
\caption{Notations and Definitions}
\begin{tabular}{cp{6.5cm}}
\toprule
Notation & Definition\\
\midrule
$\mathcal{S}$       & Condensed dataset \\
$\widetilde{\mathcal{S}}$       & Poisoned condensed dataset \\
$\mathcal{D}$    & Original dataset \\
$[\mathcal{S}/\widetilde{\mathcal{S}}/\mathcal{D}]_o$ & Subset of $\mathcal{S}/\widetilde{\mathcal{S}}/\mathcal{D}$ containing samples labeled as class $o$ \\
$o^*$ & Source class selected for crafting poisoned samples \\
$\psi$ & Downstream model \\
$f, h, \phi$ & Surrogate model $f$ of $\psi$, where $h$ is the classifier and $\phi$ is the feature extractor; $f = h \circ \phi$ \\
$\mathcal{D}_\tau^{TP}$ & Subset where samples in class $\tau$ correctly classified as $\tau$ by $f$ \\
$C$ & Set of dataset labels \\
$g_\theta$ & Attack model used to generate trigger patterns \\

% $\mathcal{T}_{triggered}$ & Triggered dataset that consists of samples from class $y_s$ that have been added adversarial triggers \\
% $\mathcal{T}_{mixed}$ & Mixture dataset that consists of clean samples from target class $y_\tau$ and triggered samples \\
\bottomrule
\end{tabular}
\label{tab:notations}
\vspace{-9pt}
\end{table}

\subsection{Threat Model}
\label{subsection: threat model}
\subsubsection{Attack Scenario}
Following the previous work~\cite{liu2023backdoor,chung2024rethinking}, we consider a scenario in which two parties collaborate on model training through data sharing. One party, possessing a large, high-quality dataset, agrees to share data with the other, who lacks access to such resources. However, due to practical constraints such as privacy concerns, proprietary limitations, or bandwidth restrictions, transferring the full dataset is infeasible. Instead, the data provider supplies a condensed version of the dataset that retains the essential characteristics of the original distribution. This is accomplished through dataset condensation, which synthesizes a small set of representative samples that preserve class semantics and learning utility. Condensed datasets are particularly valuable in settings where computational efficiency, storage constraints, or communication costs are critical. The recipient treats the condensed data as a reliable surrogate for the original dataset and incorporates it directly into the training pipeline, assuming it supports effective model learning.

In this setting, a malicious data provider can exploit their control over the condensation process to execute a backdoor attack. \textit{With full access to the original dataset and complete authority over the condensation pipeline}, the attacker generates condensed data embedded with backdoor triggers. Although the attacker does not participate in or observe the recipient’s downstream training, their influence persists through the compromised condensed dataset. These triggers induce the trained model to misbehave on specific inputs while maintaining high accuracy on standard evaluation benchmarks. This poses a significant security risk in scenarios where condensed data is assumed to be a trustworthy and safe substitute for private datasets.

\subsubsection{Attacker’s Goal}
% The attacker seeks to achieve a composite objective when embedding backdoors into condensed datasets. 
This objective involves three essential goals: stealthiness, attack effectiveness, and the preservation of clean model utility.

\paragraph{Stealthiness (STH)}
A fundamental requirement is that the injected backdoor remains stealthy. This entails two aspects. Firstly, the poisoned condensed dataset $\widetilde{\mathcal{S}}$ must be visually and statistically indistinguishable from its clean counterpart $\mathcal{S}$. This requirement is critical, as the typically small size of condensed datasets makes them particularly susceptible to anomaly detection through manual inspection or automated analysis. Secondly, test samples containing the trigger should closely resemble normal inputs in both in appearance and statistics, ensuring the attack remains inconspicuous during evaluation or deployment.

\paragraph{Attack Effectiveness}
Simultaneously, another objective of the attacker is to implant a latent backdoor that reliably activates when a specific trigger pattern is applied, while remaining dormant otherwise. Specifically, the attack effectiveness is quantified by the \textit{attack success rate (ASR)}. Let $\psi$ denote the downstream model trained on the poisoned condensed dataset $\widetilde{\mathcal{S}}$, and let $\Delta$ represent the trigger. The ASR is defined as: $ASR = \frac{1}{N_p} \sum_{i=1}^{N_p} \mathbb{I}(\psi(x_i + \Delta) = \tau)$, where $\tau$ is the target label and $N_p$ is the number of triggered test inputs. A high ASR indicates that the backdoor consistently induces misclassification to the target class.

\paragraph{Clean Model Utility}
In parallel, the attacker must ensure that the model maintains high accuracy on clean, unaltered data. This ensures that the poisoned condensed dataset remains effective for standard learning tasks and does not impair performance on benign inputs. Concretely, the clean model utility is measured by \textit{clean test accuracy (CTA)}. Let $x_i$ be a clean test sample with true label $y_i$, and $N_c$ denote the total number of such samples. The CTA is defined as: $CTA = \frac{1}{N_c} \sum_{i=1}^{N_c} \mathbb{I}(\psi(x_i) = y_i)$, where $\psi$ denote the downstream model trained on the poisoned condensed dataset $\widetilde{\mathcal{S}}$. 

\subsection{Overview of InkDrop}
InkDrop consists of three stages: (1) \textit{candidate pool construction}, (2) \textit{attack model optimization}, and (3) \textit{stealthy malicious condensation}. 

In the first stage, given a target class $\tau$, we identify a source class $o^*$ whose samples are most susceptible to being misclassified as $\tau$ by a pre-trained model. This leverages the inherent susceptibility of class $o$ to misclassification as $\tau$. Within class $o^*$, samples are ranked by their predicted confidence scores in the $\tau$-th softmax dimension, and the top $\kappa_1$ fraction is selected as candidate samples for backdoor injection.

In the second stage, an attack model is trained to generate imperceptible, instance-dependent perturbations (\textit{e.g.}, triggers) that, when applied to source samples, consistently induce misclassification into the target class. The training objective integrates four components: (i) a contrastive InfoNCE loss~\cite{oord2018representation} to align poisoned embeddings with the target class while repelling clean samples; (ii) a label alignment loss using Earth Mover’s Distance~\cite{rubner2000earth} between predicted and target soft labels; (iii) L2 regularization to limit trigger magnitude; and (iv) a learned perceptual image patch similarity loss (LPIPS)~\cite{zhang2018unreasonable} to maintain visual fidelity.

In the third stage, the trained attack model generates triggers that are applied to selected candidate samples, producing poisoned instances. These are combined with clean target-class data and fed into a dataset condensation framework. The condensation process distills this mixture into a compact synthetic dataset, embedding the backdoor functionality within the target class representation while preserving downstream task performance.

\subsection{Detailed Design of InkDrop}
\subsubsection{Candidate Pool Construction}
\begin{figure}
    \centering
    \includegraphics[width=0.7\linewidth]{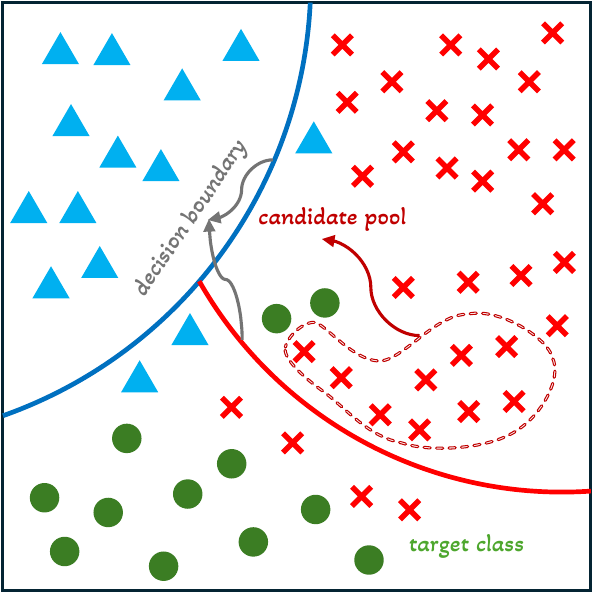}
    \caption{Illustration of constructing candidate pool.}
    \label{fig:candidate_pool}
    \vspace{-9pt}
\end{figure}

The first stage of of InkDrop involves constructing a candidate sample pool (See Figure~\ref{fig:candidate_pool}). The detailed procedure for constructing the candidate pool is presented in Algorithm~\ref{alg:cpc}. The objective is to isolate a subset of data points from the overall distribution that exhibit high ``inducibility''. Since the success of backdoor injection hinges on the perturbation's ability to cross the model's decision boundary, the selection process emphasizes samples that already exhibit boundary ambiguity with the target class. Applying triggers indiscriminately across the dataset not only reduces injection efficiency but also increases the risk of detection by anomaly detectors. Therefore, constructing a candidate pool that responds sensitively to perturbations is a critical foundation for effective and stealthy attacks.

\paragraph{Source Class Selection}
The first step in candidate pool construction involves selecting the most attack-susceptible source class $o^*$ from all non-target classes $o \ne \tau$. The strategy is grounded in analyzing the model’s misclassification patterns through the lenses of information flow and decision uncertainty. By exploiting inter-class information confusion, this step establishes the most favorable pathway for effective backdoor injection.

To quantify the informational proximity between the source and target classes, we measure the posterior shift between the source class distribution and the target class representation. This process involves two steps: (i) constructing a reference distribution for the target class, and (ii) computing the information shift using Kullback–Leibler (KL) divergence.

(i) reference distribution construction. Firstly, we construct a reference output distribution $q_\tau$ for the target class $\tau$, defined as the model’s average predictive distribution over true positive samples within the target class $\tau$. Let $f(x)$ denote the model’s predicted probability for an input $x$. Then,
\begin{equation}
\begin{aligned}
    q_\tau(y)&\triangleq\mathbb{E}_{x\sim\mathcal{D}_\tau^{TP}}[f(y|x)]\\
    \mathcal{D}_\tau^{\mathrm{TP}} &= \left\{ x \in \mathcal{D}_\tau \mid \arg\max f(x) = \tau \right\}
\end{aligned}
\end{equation}
where $\mathcal{D}_\tau^{TP}$ denotes the subset of target-class samples correctly classified as $\tau$. This formulation excludes low-confidence samples near the decision boundary and captures a stable, high-confidence representation of the target class in the model's output space.

(ii) information shift measurement. Secondly, for any sample $x$, we quantify its \textit{information shift} toward the target class $\tau$ by computing the KL divergence between its predictive distribution $f(x)$ and the target class reference distribution $q_\tau$:
\begin{equation}
    D_{\mathrm{KL}}(f(x) \parallel q_\tau) = \sum_{c=1}^C f_c(x) \log \frac{f_c(x)}{q_\tau(c)}
\end{equation}
where $C$ represents the number of the classes in the dataset. Smaller $D_{\mathrm{KL}}(f(x) \parallel q_\tau)$ indicates that the sample’s prediction aligns more closely with the model’s typical output for class $\tau$, implying an inherent bias toward the target class $\tau$ in the model's decision space.

\begin{algorithm}[tb]
\renewcommand{\algorithmicrequire}{\textbf{Input:}}
\renewcommand{\algorithmicensure}{\textbf{Output:}}
\caption{Candidate Pool Construction}
\label{alg:cpc}
\begin{algorithmic}[1]
\REQUIRE Original data $\mathcal{D}$, target class $\tau$, label set $C$, hyper-parameters $\lambda$, $\kappa_1$, $\kappa_2$
\ENSURE Candidate pool $\mathcal{P}_\tau$, 
\STATE Train the surrogate model $f$ on $\mathcal{D}$
\STATE Construct $\mathcal{D}_\tau^{TP}$ as the set of samples in class $\tau$ correctly classified by $f$
\STATE Construct $\mathcal{T}_\tau^{\kappa_2} = \text{Top-}\kappa_2\left( \mathcal{D}_\tau^{TP}; f_\tau(x) \right)$, with $|\mathcal{T}_\tau^{\kappa_2}| = m$
\STATE Estimate the reference distribution: $q_\tau \approx \frac{1}{m} \sum_{j=1}^m f(y \mid x_j), \quad x_j \in \mathcal{T}_\tau^{\kappa_2}$
\STATE Initialize $\Gamma = \varnothing$
\FOR{each $o \in C$}
    \STATE $\hat{\mathcal{I}}_{o \rightarrow \tau} = \frac{1}{n} \sum_{i=1}^n D_{\mathrm{KL}}(f(x_i) \parallel q_\tau)\quad$// Informational proximity
    \STATE $\hat{\pi}_{o \rightarrow \tau} = \frac{1}{n} \sum_{i=1}^n \mathbb{I} \left[ \arg\max f(x_i) = \tau \right]\quad$// Misclassification rate
    \STATE $e_o = \hat{\mathcal{I}}_{o \rightarrow \tau} - \lambda \hat{\pi}_{o \rightarrow \tau}\quad$// Joint criterion score
    \STATE Add $e_o$ to $\Gamma$
\ENDFOR
\STATE $o^* \leftarrow \arg\min_o \Gamma\quad$// Select the source class with minimal score
\STATE $\mathcal{P}_\tau = \left\{ x_i \in \mathcal{D}_{o^*} \mid \text{rank}(f_\tau(x_i)) \leq \kappa_1 \cdot |\mathcal{D}_{o^*}| \right\}\quad$// Construct candidate pool
\RETURN $\mathcal{P}_\tau$
\end{algorithmic}
\end{algorithm}

To assess the overall informational proximity of a source class $o$ to the target class $\tau$, we compute the expected KL divergence over the source class distribution:
\begin{equation}
    \mathcal{I}_{o \rightarrow \tau} \triangleq \mathbb{E}_{x \sim \mathcal{D}_o} \left[ D_{\mathrm{KL}}(f(x) \parallel q_\tau) \right]
\end{equation}
where $\mathcal{D}_o$ represents the source class data. 

This expectation reflects the posterior information closeness of class $o$ to $\tau$. Lower values indicate that the source class is more easily absorbed into the target class distribution, highlighting a more exploitable pathway for backdoor injection.

To further characterize the misclassification tendency of a source class $o$ toward the target class $\tau$, we define the misclassification rate as:
\begin{equation}
    \pi_{o \rightarrow \tau} = \mathbb{P}_{x \sim \mathcal{D}_o} \left[ \arg\max f(x) = \tau \right]
\end{equation}
This metric captures the probability that samples from $\mathcal{D}_o$ are classified as $\tau$, quantifying the extent to which class $o$ intrudes into the decision region of the target class $\tau$. 

Finally, we unify posterior information proximity and misclassification probability into a joint selection criterion:
\begin{equation}
    o^* = \arg\min_{o \ne \tau} \left( \mathcal{I}_{o \rightarrow \tau} - \lambda \pi_{o \rightarrow \tau} \right)
\end{equation}
Here, $\lambda$ is a trade-off parameter that balances structural alignment with the target class and the ability to penetrate the decision boundary. This criterion identifies the most vulnerable inter-class pathway within the model, offering both theoretical grounding and structural guidance for downstream attack sample construction.

Since the expectations above are not analytically tractable, we adopt Monte Carlo estimation in practice:

Reference Distribution $q_\tau$: We select true positive samples from the target class that are predicted as $\tau$, then retain the top-$\kappa_2$ samples ranked by confidence in class $\tau$. Averaging their softmax outputs yields the approximation:
\begin{equation}
\begin{aligned}
    q_\tau(y) &\approx \frac{1}{m} \sum_{j=1}^m f(y \mid x_j), \quad x_j \in \mathcal{T}_\tau^{\kappa_2}\\
    \mathcal{T}_\tau^{\kappa_2}& = \text{Top-}\kappa_2\left( \mathcal{D}_\tau^{\mathrm{TP}}; f_\tau(x) \right)
\end{aligned}
\end{equation}
where $\text{Top-}\kappa_2(\cdot)$ denotes selecting the top $\kappa_2$ proportion ranked by $f_\tau(x)$ in descending order.

Information Proximity $\mathcal{I}_{o \rightarrow \tau}$: We draw $n$ samples from the source class $\mathcal{D}_o$ and estimate the expected KL divergence:

\begin{equation}
    \hat{\mathcal{I}}_{o \rightarrow \tau} = \frac{1}{n} \sum_{i=1}^n D_{\mathrm{KL}}(f(x_i) \parallel q_\tau)
\end{equation}

Misclassification Rate $\pi_{o \rightarrow \tau}$: This is estimated using an indicator function:

\begin{equation}
    \hat{\pi}_{o \rightarrow \tau} = \frac{1}{n} \sum_{i=1}^n \mathbb{I} \left[ \arg\max f(x_i) = \tau \right]
\end{equation}

These three estimators constitute a practical and theoretically grounded mechanism for source class selection, designed to ensure structural optimality for backdoor injection.

Moreover, in Subsection~\ref{subsection: threat model}, we assume the attacker lacks direct access to the downstream model. In practice, a surrogate model $f(\cdot)$ is trained on accessible data to approximate the downstream model’s decision behavior. This model is trained conventionally on the original dataset $\mathcal{D}$ and captures, to some extent, the downstream model’s class confusion patterns and decision boundary structure. All subsequent analysis and sample construction are performed exclusively on this surrogate. Despite lacking access to the downstream model’s parameters and training specifics, carefully crafted strategies on the surrogate can produce transferable attack effects.

\paragraph{Candidate Attack Sample Selection}
After determining the source class $o^*$, we further extract the most attack-prone subset from its sample set $\mathcal{D}_{o^*}$ to form the final candidate pool. The goal is to identify samples that naturally reside near the decision boundary of the target class $\tau$, making them more susceptible to misclassification under minimal perturbation.

To identify optimal attack candidates, we use the model’s output on the target class dimension, $f_\tau(x)$, as a confidence score indicating the likelihood assigned to class $\tau$. For each sample $x \in \mathcal{D}_{o^*}$, we define its target-class confidence as $\zeta= f_\tau(x)$. Higher values of $\zeta$ suggest that the model already exhibits a strong bias toward the target class, characterized by low predictive uncertainty and feature representations concentrated near the center of the target class distribution. These samples are located near the decision boundary but skewed toward the target class, making them prime candidates for backdoor injection. Minimal perturbations are sufficient to alter their predicted label, facilitating effective attack transfer.

Based on the above analysis, we rank all samples in $\mathcal{D}_{o^*}$ in descending order of $f_\tau(x_i)$, and select the top $\kappa_1$ quantile to construct the candidate pool $\mathcal{P}_\tau$:

\begin{equation}
    \mathcal{P}_\tau = \left\{ x_i \in \mathcal{D}_{o^*} \mid \text{rank}(f_\tau(x_i)) \leq \kappa_1 \cdot |\mathcal{D}_{o^*}| \right\}
\end{equation}

This procedure serves as a form of posterior entropy compression, favoring samples where the model exhibits a strong discriminative tendency toward the target class. Compared to low-confidence instances, these samples are more susceptible to perturbation, more likely to trigger misclassification, and require minimal injection effort.

The resulting candidate pool $\mathcal{P}_\tau$ forms a subset that is naturally close to the target class, providing a stable and efficient medium for backdoor trigger injection. 

\begin{algorithm}[tb]
\renewcommand{\algorithmicrequire}{\textbf{Input:}}
\renewcommand{\algorithmicensure}{\textbf{Output:}}
\caption{Attack Model Training}
\label{alg:amt}
\begin{algorithmic}[1]
\REQUIRE Candidate pool $\mathcal{P}_\tau$, $\mathcal{T}_\tau^{\kappa_2}$, untrained attack model $g_\theta$, hyper-parameters $\lambda_1,\lambda_2,\lambda_3,\lambda_4$
\ENSURE Trained attack model $g_{\theta}$
\STATE $\mathcal{Z}_\tau = \left\{ \phi(x) \mid x \in \mathcal{T}_\tau^{\kappa_2} \right\}\quad$// Target embedding set
\STATE $\mathcal{Q}_\tau = \left\{ f(x) \mid x \in \mathcal{T}_\tau^{\kappa_2} \right\}\quad$// Target soft label set
\FOR{epoch $e$ from 1 to $E$}
\STATE Compute a feature-space contrastive loss $\mathcal{L}_{contrast}$ using Eq (\ref{eq:contrast})
\STATE Compute a label-space alignment loss $\mathcal{L}_{soft}$ using Eq (\ref{eq:soft})
\STATE Compute a perturbation regularization term using Eq (\ref{eq:L2})
\STATE Compute a perceptual consistency loss $\mathcal{L}_{LPIPS}$ using Eq (\ref{eq:lpips})
\STATE $\mathcal{L}=\lambda_1 \cdot \mathcal{L}_{\mathrm{contrast}} + \lambda_2 \cdot \mathcal{L}_{\mathrm{soft}} + \lambda_3 \cdot \mathcal{L}_{\mathrm{L2}} + \lambda_4 \cdot \mathcal{L}_{\mathrm{LPIPS}}$
\STATE Update parameters of $g_\theta$ using $\mathcal{L}$
\ENDFOR
\RETURN $g_\theta$
\end{algorithmic}
\end{algorithm}

\subsubsection{Attack Model Training}
To enable imperceptible and controllable backdoor injection, we train an attack model $g_\theta$ that generates instance-dependent perturbations for source samples while maintaining visual fidelity. The objective is to induce misclassification into a designated target class $\tau$ under the surrogate model. The attack model learns a perturbation mapping from the input space to the posterior output space, optimizing structural alignment and discriminative transfer to facilitate effective class induction. The training procedure for optimizing the attack model is detailed in Algorithm~\ref{alg:amt}.

For clarity, we represent the surrogate model $f(\cdot)$ as a composition of a feature extractor and a classifier, that is, $f = h \circ \phi$, where $\phi(\cdot)$ denotes the feature extractor and $h(\cdot)$ the classifier. The model’s prediction for an input image $x$ is expressed as:
\begin{equation}
    f(x) = h(\phi(x)) \in \Delta^C
\end{equation}

where $\Delta^C$ denotes the probability simplex over $C$ classes.

To extract representative structural information for the target class $\tau$,  we compute the feature representation $\phi(x)$ and model output $f(x)$ for each sample $x \in \mathcal{T}_\tau^{\kappa_2}$, forming two reference sets:

Target embedding set:

\begin{equation}
    \mathcal{Z}_\tau = \left\{ \phi(x) \mid x \in \mathcal{T}_\tau^{\kappa_2} \right\}
\end{equation}

Target soft label set:

\begin{equation}
    \mathcal{Q}_\tau = \left\{ f(x) \mid x \in \mathcal{T}_\tau^{\kappa_2} \right\}
\end{equation}

These sets serve as supervisory signals during the training of the attack model $g_\theta$, guiding poisoned samples to converge toward the target class in both the feature and output spaces via contrastive learning and label alignment.

Subsequently, leveraging the previously constructed target-class reference sets $\mathcal{Z}_\tau$ and $\mathcal{Q}_\tau$, we train an attack model $g_\theta$ to learn a instance-dependent perturbation strategy. Given a source-class input $x \in \mathcal{D}_{o^*}$, the model generates a perturbation $\delta = g_\theta(x)$, yielding a poisoned sample $x^{\mathrm{atk}} = x + \delta$.

The perturbation must satisfy two key constraints: it should (1) reliably induce misclassification into the target class $\tau$, ensuring target alignment in both representation and output space, and (2) remain imperceptible and structurally consistent to avoid triggering detection mechanisms or visual anomalies. To achieve both objectives, we design a multi-component loss function to optimize $g_\theta$, integrating four complementary terms:
\begin{itemize}
    \item
    Feature-space contrastive loss (InfoNCE): Encourages the representation $\phi(x^{\mathrm{atk}})$ to align with embeddings in $\mathcal{Z}_\tau$ while remaining distant from clean features.

    \item
    Label-space alignment loss (EMD): Aligns the model output $f(x^{\mathrm{atk}})$ with the soft label set $\mathcal{P}_\tau$.

    \item
    Perturbation regularization ($L_2$): Penalizes the norm of $\delta$ to constrain perturbation magnitude.

    \item
    Perceptual consistency loss (LPIPS): Maintains visual fidelity by minimizing perceptual distance between $x$ and $x^{\mathrm{atk}}$.
\end{itemize}

\paragraph{Feature-space contrastive loss (InfoNCE)}
To guide poisoned samples toward the target class in feature space while repelling them from the source class, we incorporate a contrastive loss based on the InfoNCE formulation. This objective maximizes similarity between poisoned and target-class embeddings while minimizing similarity with embeddings from clean source-class samples.

Let $\{x_i^{\mathrm{atk}}\}_{i=1}^B$ denote a batch of poisoned samples generated by the attack model, with normalized feature embeddings $\{\hat{z}_i^{\mathrm{atk}}=\phi(x_i^{atk})\}_{i=1}^B$. For each $x_i^{\mathrm{atk}}$, we sample a positive embedding $\hat{z}_i^{\mathrm{target}}$ from true positive target-class instances (i.e., $\mathcal{Z}_\tau$), and its corresponding negative embeddings $\{\hat{z}_j^{\mathrm{clean}}=\phi(x_j)\}_{j=1}^B$. The contrastive loss for the $i$-th sample is defined as:

\begin{equation}
    \mathcal{L}_{contrast} = -\log\frac{\exp\left(\langle \hat{z}_i^{\mathrm{atk}}, \hat{z}_i^{\mathrm{target}} \rangle / \nu\right)}{\sum_{j=1}^B \exp\left(\langle \hat{z}_i^{\mathrm{atk}}, \hat{z}_j^{\mathrm{clean}} \rangle / \nu\right)}
\label{eq:contrast}
\end{equation}
Here, $\nu$ is a temperature parameter that controls distribution sharpness. All embeddings are normalized to unit norm, making the inner product equivalent to cosine similarity.

\paragraph{Label-space alignment loss}
To improve the prediction consistency of poisoned samples toward the target class, we introduce a label-space alignment objective based on Earth Mover’s Distance (EMD). This metric quantifies the discrepancy between the output distribution of a poisoned sample and that of a high-confidence target-class sample under the surrogate model.

During training, we select samples from $\mathcal{T}_\tau^{\kappa_2}$ and extract their classifier outputs $\{q_i\}$ as target soft labels. Concurrently, we sample source-class candidates, apply perturbations from the attack model to obtain poisoned samples $\{x_i^{\mathrm{atk}}\}$, and collect their outputs $\{p_i\}$.

The EMD loss for each pair $(p_i, q_i)$ is computed as:

\begin{equation}
    \mathcal{L}_{soft} = \min_{T \in \Pi(p_i, q_i)} \sum_{c_1, c_2} T(c_1, c_2) \cdot d(c_1, c_2)
\label{eq:soft}
\end{equation}
Here, $T$ is the transport matrix, $d(c_1, c_2)$ denotes the ground distance between labels, and $\Pi(p_i, q_i)$ is the set of transport plans satisfying marginal constraints.

\paragraph{Perturbation Regularization}
To constrain the perturbation magnitude in image space and reduce the risk of detection by anomaly detectors or human observers, we incorporate an $L_2$ regularization term on the trigger generated by the attack model. Let $\delta_i = g_\theta(x_i)$ denote the perturbation for each input sample. The regularization loss is defined as:
\begin{equation}
    \mathcal{L}_{\mathrm{L2}} = \frac{1}{B} \sum_{i=1}^B \| \delta_i \|_2^2
\label{eq:L2}
\end{equation}
This term encourages the perturbations to be minimal in norm, promoting sparsity and locality, which enhances the imperceptibility and stealth of the attack.

\paragraph{Perceptual Consistency Loss}
While the $L_2$ regularization term constrains perturbation magnitude in Euclidean space, it fails to capture perceptual sensitivity to structural and semantic changes in images. Therefore, we incorporate a structure-aware constraint using the Learned Perceptual Image Patch Similarity (LPIPS) loss, which promotes consistency between poisoned and clean images in the perceptual space. LPIPS measures structural discrepancies by comparing normalized activations from multiple deep feature layers, offering finer sensitivity to semantic and textural differences than pixel-based metrics.

Let $\hat{y}^l$ and $\hat{y}_{0hw}^l$ denote the normalized output of the original and poisoned images from the $l$-th layer of a pre-trained perceptual network, with spatial dimensions $H_l$ and $W_l$, and let $w_l$ denote channel-wise weighting. For a pair of original and poisoned images $(x_i, x_i^{\mathrm{atk}})$, the LPIPS loss is defined as:
\begin{equation}
    \mathcal{L}_{\mathrm{LPIPS}}(x_i^{\mathrm{atk}}, x_i) = \sum_l \frac{1}{H_l W_l} \sum_{h,w} \left\| w_l \odot \left( \hat{y}^l_{hw} - \hat{y}^l_{0hw} \right) \right\|_2^2
\label{eq:lpips}
\end{equation}
where $\odot$ denotes element-wise (Hadamard) multiplication across channels, and the differences are averaged over spatial dimensions. The final loss is computed as the batch mean across all samples. This constraint mitigates visually perceptible distortions, enhancing both the stealth and perceptual integrity of the injected backdoor. The overall training objective combines four complementary loss terms:
\begin{equation}
    \mathcal{L} = \lambda_1 \cdot \mathcal{L}_{\mathrm{contrast}} + \lambda_2 \cdot \mathcal{L}_{\mathrm{soft}} + \lambda_3 \cdot \mathcal{L}_{\mathrm{L2}} + \lambda_4 \cdot \mathcal{L}_{\mathrm{LPIPS}}
\end{equation}
where $\lambda_1, \lambda_2, \lambda_3, \lambda_4$ are scalar weights that balance the contributions of contrastive loss, soft label alignment, perturbation magnitude control, and perceptual consistency, respectively.

\subsubsection{Stealthy Malicious Condensation}
Once the attack model $g_\theta$ is trained to generate perturbations that induce targeted misclassification, we integrate the resulting adversarial examples into the condensation pipeline. Specifically, the condensation process for the target class is modified to include triggered samples, ensuring that the backdoor behavior is encoded in the surrogate model and subsequently transferred to the downstream model.

\begin{figure*}[t]
    \centering

    \begin{minipage}[b]{0.497\linewidth}
        \centering
        \includegraphics[width=\linewidth]{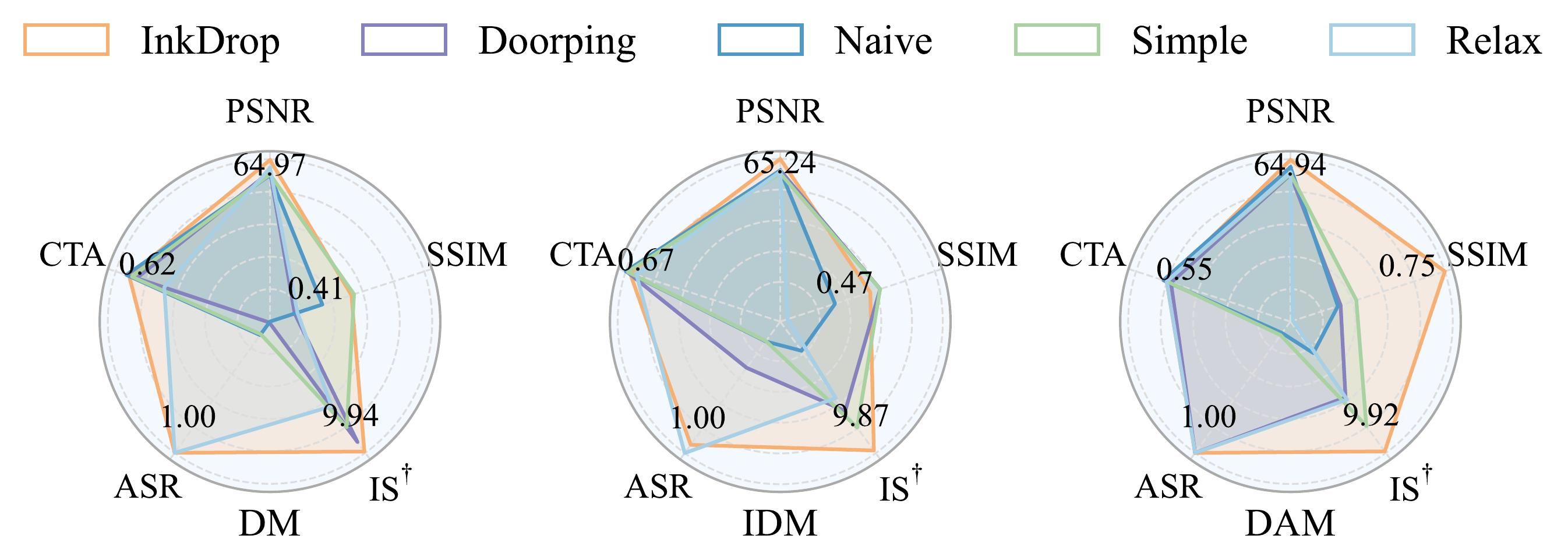}
        (a) STL-10
    \end{minipage}
    \hfill
    \begin{minipage}[b]{0.497\linewidth}
        \centering
        \includegraphics[width=\linewidth]{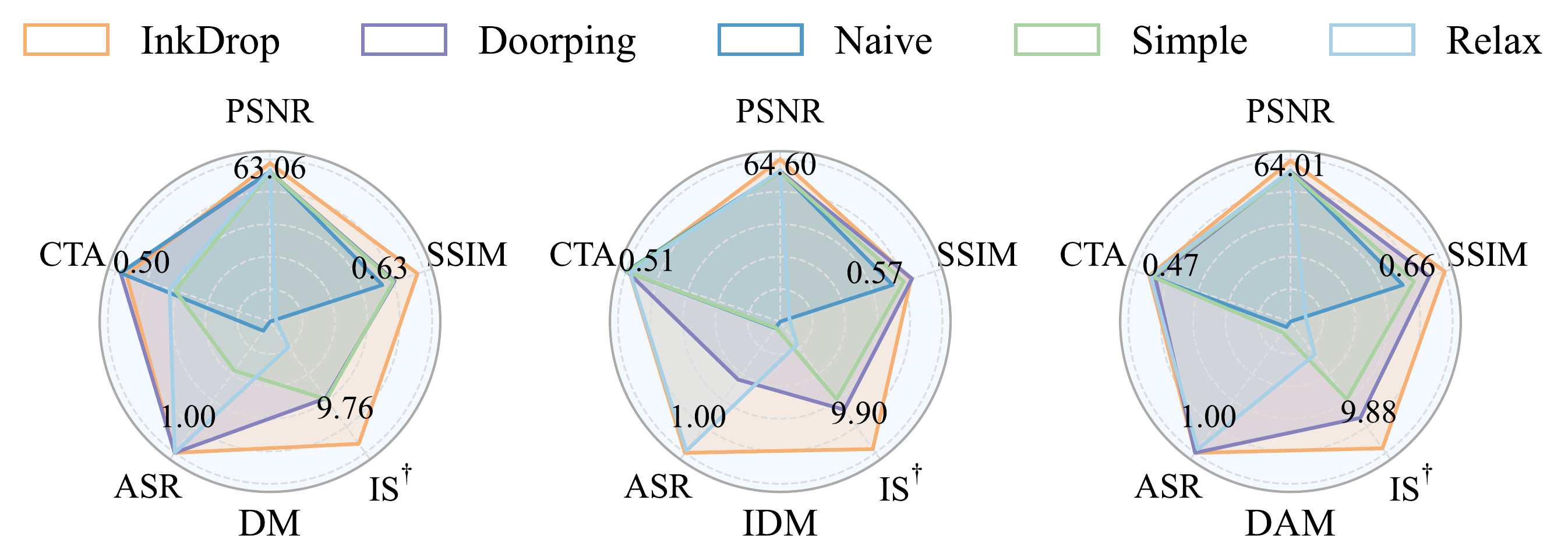}
        (b) Tiny-ImageNet
    \end{minipage}

    \begin{minipage}[b]{0.497\linewidth}
        \centering
        \includegraphics[width=\linewidth]{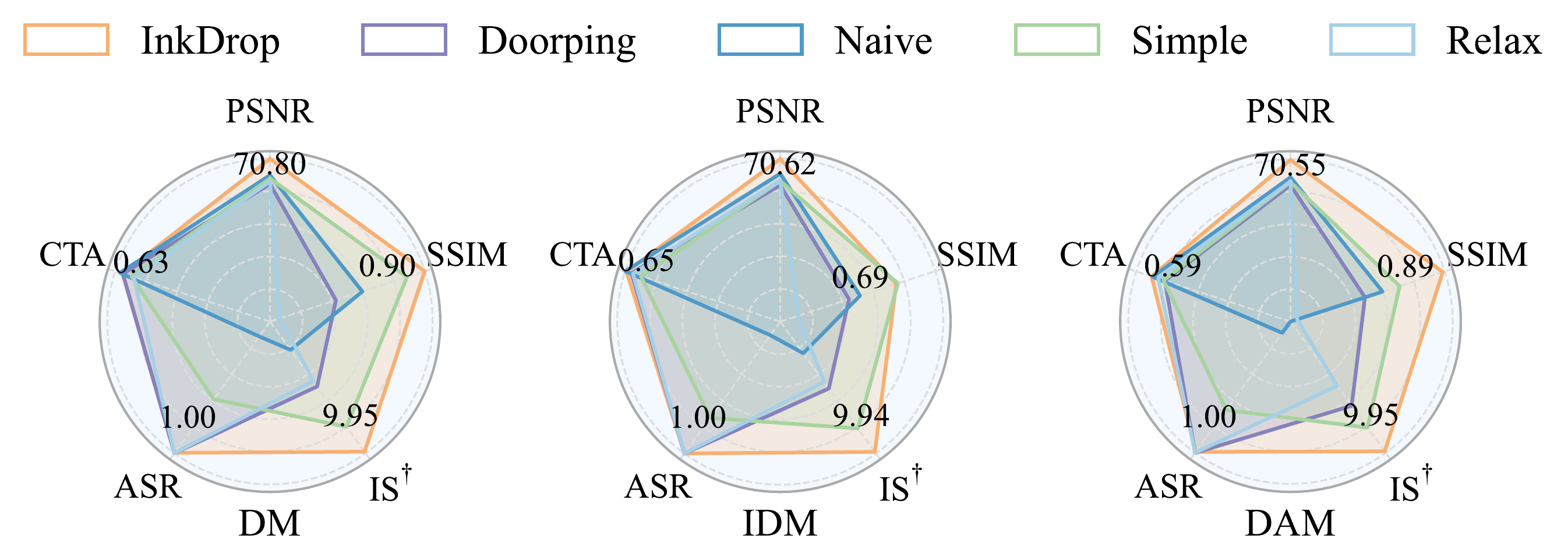}
        (c) CIFAR-10
    \end{minipage}
    \hfill
    \begin{minipage}[b]{0.497\linewidth}
        \centering
        \includegraphics[width=\linewidth]{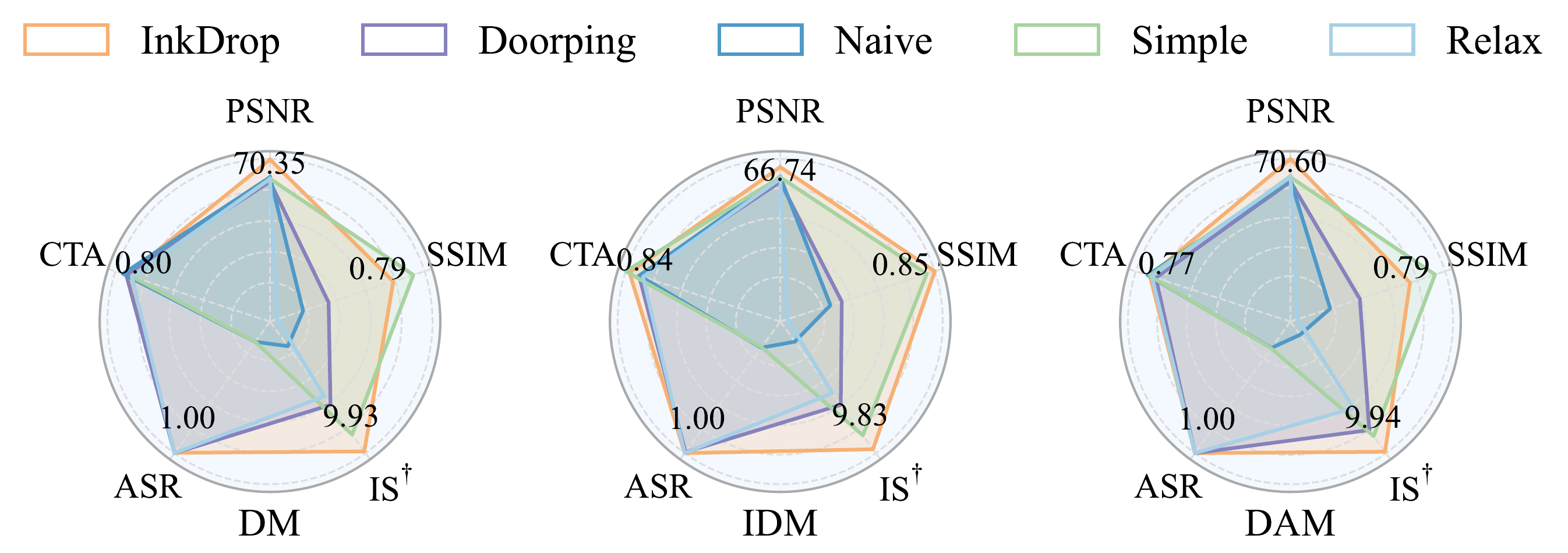}
        (d) SVHN
    \end{minipage}

    \caption{
    Radar plots showing normalized performance across attack success rate (ASR), clean task accuracy (CTA), and stealthiness (STE) for each method. Each subplot corresponds to a different dataset. Curves extending toward the outer edge indicate stronger overall performance.
    }
    \label{fig:radar}
\vspace{-9pt}
\end{figure*}

\paragraph{Constructing the Mixed Data}
To embed backdoor behavior into the condensation data, we begin by constructing a mixed dataset that combines both clean and triggered samples. Given the candidate pool $\mathcal{P}_\tau$, a set of corresponding trigger patterns $\Delta\triangleq\{\delta_i=g_\theta(x_i)\mid x_i\in\mathcal{P}_\tau\}_{i=1}^{\kappa_1\cdot|\mathcal{D}_{o^*}|}$ is generated by the attack model $g_\theta$. Each pattern $\delta_i$ is applied to the corresponding sample $x_i \in \mathcal{P}_\tau$, yielding the triggered set:
\begin{equation}
    \mathcal{T}_\text{trigger}=\{x_i+\delta_i\mid x_i\in\mathcal{P}_\tau,\delta_i\in\Delta\}
\end{equation}
To avoid overwhelming the original class distribution while still injecting the backdoor signal, only a small fraction of the triggered samples is incorporated into the clean target class. Specifically, we randomly select $\rho \cdot\kappa_1\cdot|\mathcal{D}_{o^*}|$ samples from $\mathcal{T}_\text{trigger}$ and merge them with the clean samples from $\mathcal{D}\tau$. The resulting mixed dataset is defined as:
\begin{equation}
\mathcal{T}_{\text{mixed}} = \mathcal{D}_\tau \cup \{ ({x}_i^{atk}, \tau) \}_{i=1}^{\rho \cdot\kappa_1\cdot|\mathcal{D}_{o^*}|},\quad{x}_i^{atk}\in\mathcal{T}_\text{trigger}
\end{equation}
This blending mechanism introduces the trigger pattern into the target class distribution in a controlled manner. It biases the model toward associating the trigger with class $\tau$ while preserving generalization to clean inputs.

\paragraph{Malicious Condensation}
In the next step, we synthesize a condensed dataset $\widetilde{\mathcal{S}}$ that preserves model utility while embedding the intended backdoor behavior. For each class $o \ne \tau$, standard dataset condensation is performed using only the clean data $\mathcal{D}_o$. The objective is to match the feature distributions of the real and synthetic samples:
\begin{equation}
\begin{aligned}
    \mathcal{S}_o = &\underset{\mathcal{S}_o}{\arg\min} \ \mathbb{E}_{x \sim p_{\mathcal{D}_o}, x' \sim p_{\mathcal{S}_o}, \theta \sim p_\theta} \\
    &D\left(P_{\mathcal{D}_o}(x; \theta), P_{\mathcal{S}_o}(x'; \theta)\right) + \lambda \mathcal{R}(\mathcal{S}_o)
\end{aligned}
\end{equation}
$P_{\mathcal{D}_o}(x; \theta)$ and $P{\mathcal{S}_o}(x'; \theta)$ denote the model-induced feature distributions over real and synthetic samples, respectively, and $\mathcal{R}(\mathcal{S}_o)$ is a regularization term.

For the target class $\tau$, condensation is carried out using the mixed dataset $\mathcal{T}_{\text{mixed}}$ constructed in the previous step. This allows the resulting synthetic subset $\widetilde{\mathcal{S}}_\tau$ to capture both the clean semantics and the adversarial characteristics introduced via the trigger. The optimization objective is:
\begin{equation}
\begin{aligned}
    \widetilde{\mathcal{S}}\tau = &\underset{\widetilde{\mathcal{S}}\tau}{\arg\min} \ \mathbb{E}_{x \sim p_{\mathcal{T}_{\text{mixed}}}, x' \sim p_{\widetilde{\mathcal{S}}_\tau}, \theta \sim p_\theta} \\
    &D\left(P_{\mathcal{T}_{\text{mixed}}}(x; \theta), P_{\widetilde{\mathcal{S}}_\tau}(x'; \theta)\right) + \lambda \mathcal{R}(\widetilde{\mathcal{S}}_\tau)
\end{aligned}
\end{equation}

The final condensed dataset is assembled as $\widetilde{\mathcal{S}} = \bigcup_{o \ne \tau} \mathcal{S}_o \cup \widetilde{\mathcal{S}}\tau$, combining clean synthetic subsets for non-target classes with the backdoor-aware condensed subset for the target class.

\section{Experiments}

\subsection{Dataset and Networks}
We evaluate InkDrop on four standard image classification benchmarks: STL-10~\cite{coates2011analysis}, Tiny-ImageNet~\cite{le2015tiny}, CIFAR-10~\cite{krizhevsky2009learning}, and SVHN~\cite{netzer2011reading}. These datasets cover a wide spectrum of visual complexity, semantic granularity, and image resolution, offering a robust basis for assessing the generalizability of InkDrop. Specifically, we retain 50 synthetic samples per class. To construct the synthetic data, we employ two commonly adopted architectures: \textit{ConvNet}, representing a compact model, and \textit{AlexNetBN}~\cite{NIPS2012_c399862d}, which represent lightweight and moderately expressive condensation encoders. For downstream model training and evaluation, we utilize four network architectures of increasing complexity: ConvNet, AlexNetBN, VGG11~\cite{simonyan2014very}, and ResNet18~\cite{he2016deep}. We compare InkDrop against four recent state-of-the-art backdoor injection methods: \textit{Naive}\cite{liu2023backdoor}, \textit{Doorping}\cite{liu2023backdoor}, \textit{Simple}\cite{chung2024rethinking}, and \textit{Relax}\cite{chung2024rethinking}, enabling a comprehensive evaluation of its effectiveness across diverse settings.

% --- 颜色定义 (您可以根据喜好修改) ---
\definecolor{highlightcolor}{rgb}{0.88, 0.92, 1.0} % 定义您方法的高亮颜色 (淡蓝色)
% 表头分区颜色
\definecolor{dmcolor}{rgb}{0.85, 0.95, 0.85}       % 为 DM 定义淡绿色
\definecolor{idmcolor}{rgb}{1.0, 0.95, 0.85}      % 为 IDM 定义淡橙色
\definecolor{damcolor}{rgb}{0.9, 0.9, 1.0}        % 为 DAM 定义淡紫色

\begin{table*}[ht]
\centering
\caption{Attack performance with ConvNet as the condensation backbone.}
\label{tab:convnet_final}
\begin{tabular}{clrrrrrr} 
\toprule
% ------- 第一行表头 -------
\multicolumn{2}{>{\columncolor{gray!20}}c}{} &
\multicolumn{2}{>{\columncolor{dmcolor}}c}{\textbf{DM}} &
\multicolumn{2}{>{\columncolor{idmcolor}}c}{\textbf{IDM}} &
\multicolumn{2}{>{\columncolor{damcolor}}c}{\textbf{DAM}} \\

% ------- 第二行表头（负行数 multirow） -------
\multirow{-2}{*}{\cellcolor{gray!20}\textbf{Dataset}} &
\multirow{-2}{*}{\cellcolor{gray!20}\textbf{Method}} &
\multicolumn{1}{>{\columncolor{dmcolor}}c}{\textbf{CTA}} &
\multicolumn{1}{>{\columncolor{dmcolor}}c}{\textbf{ASR}} &
\multicolumn{1}{>{\columncolor{idmcolor}}c}{\textbf{CTA}} &
\multicolumn{1}{>{\columncolor{idmcolor}}c}{\textbf{ASR}} &
\multicolumn{1}{>{\columncolor{damcolor}}c}{\textbf{CTA}} &
\multicolumn{1}{>{\columncolor{damcolor}}c}{\textbf{ASR}} \\
\midrule

% --- 数据集: STL10 ---
\multirow{6}{*}{\textbf{STL10}}
& \cellcolor{highlightcolor}InkDrop   & \cellcolor{highlightcolor}$\result{0.5805}{0.0008}$ & \cellcolor{highlightcolor}$\result{1.0000}{0.0000}$ & \cellcolor{highlightcolor}$\result{0.6476}{0.0007}$ & \cellcolor{highlightcolor}$\result{0.9333}{0.0000}$ & \cellcolor{highlightcolor}$\result{0.5350}{0.0011}$ & \cellcolor{highlightcolor}$\result{1.0000}{0.0000}$ \\
% & Sneakdoor  & $\result{0.5979}{0.0006}$ & $\result{0.9725}{0.0000}$ & $\result{0.6582}{0.0005}$ & $\result{0.9790}{0.0009}$ & $\result{0.5324}{0.0007}$ & $\result{0.9918}{0.0006}$ \\
& Doorping     & $\result{0.5773}{0.0008}$ & $\result{0.1488}{0.0069}$ & $\result{0.6609}{0.0010}$ & $\result{0.3144}{0.0153}$ & $\result{0.5328}{0.0008}$ & $\result{1.0000}{0.0000}$ \\
& Naive    & $\result{0.6210}{0.0006}$ & $\result{0.1032}{0.0060}$ & $\result{0.6671}{0.0005}$ & $\result{0.1024}{0.0073}$ & $\result{0.5490}{0.0006}$ & $\result{0.0876}{0.0085}$ \\
& Simple     & $\result{0.5972}{0.0005}$ & $\result{0.0964}{0.0094}$ & $\result{0.6576}{0.0005}$ & $\result{0.1000}{0.0068}$ & $\result{0.5346}{0.0007}$ & $\result{0.1032}{0.0037}$ \\
& Relax     & $\result{0.5962}{0.0009}$ & $\result{0.9996}{0.0008}$ & $\result{0.6582}{0.0008}$ & $\result{0.9536}{0.0110}$ & $\result{0.5348}{0.0011}$ & $\result{1.0000}{0.0000}$ \\
\addlinespace

% --- 数据集: TINY IMAGENET ---
\multirow{6}{*}{\shortstack{\textbf{Tiny}\\\textbf{ImageNet}}}
& \cellcolor{highlightcolor}InkDrop   & \cellcolor{highlightcolor}$\result{0.4764}{0.0024}$ & \cellcolor{highlightcolor}$\result{1.0000}{0.0000}$ & \cellcolor{highlightcolor}$\result{0.5000}{0.0047}$ & \cellcolor{highlightcolor}$\result{1.0000}{0.0000}$ & \cellcolor{highlightcolor}$\result{0.4688}{0.0037}$ & \cellcolor{highlightcolor}$\result{1.0000}{0.0000}$ \\

% & Sneakdoor  & $\result{0.5026}{0.0010}$ & $\result{1.0000}{0.0000}$ & $\result{0.5174}{0.0036}$ & $\result{1.0000}{0.0000}$ & $\result{0.4822}{0.0032}$ & $\result{1.0000}{0.0000}$ \\
& Doorping     & $\result{0.4958}{0.0023}$ & $\result{1.0000}{0.0000}$ & $\result{0.5120}{0.0048}$ & $\result{1.0000}{0.0000}$ & $\result{0.4490}{0.0030}$ & $\result{1.0000}{0.0000}$ \\
& Naive    & $\result{0.4968}{0.0020}$ & $\result{0.0704}{0.0020}$ & $\result{0.5014}{0.0075}$ & $\result{0.0424}{0.0038}$ & $\result{0.4620}{0.0029}$ & $\result{0.0418}{0.0019}$ \\
& Simple     & $\result{0.4932}{0.0032}$ & $\result{0.1000}{0.0038}$ & $\result{0.5086}{0.0028}$ & $\result{0.0462}{0.0015}$ & $\result{0.4582}{0.0030}$ & $\result{0.0820}{0.0024}$ \\
& Relax     & $\result{0.4942}{0.0029}$ & $\result{0.9960}{0.0000}$ & $\result{0.4840}{0.0056}$ & $\result{0.9412}{0.0019}$ & $\result{0.4646}{0.0019}$ & $\result{0.9730}{0.0011}$ \\
\addlinespace

% --- 数据集: CIFAR10 ---
\multirow{6}{*}{\textbf{CIFAR10}}
% --- 使用 \cellcolor 高亮您的 InkDrop 方法 ---
& \cellcolor{highlightcolor}InkDrop   & \cellcolor{highlightcolor}$\result{0.6217}{0.0006}$ & \cellcolor{highlightcolor}$\result{0.9967}{0.0000}$ & \cellcolor{highlightcolor}$\result{0.6471}{0.0009}$ & \cellcolor{highlightcolor}$\result{1.0000}{0.0000}$ & \cellcolor{highlightcolor}$\result{0.5872}{0.0005}$ & \cellcolor{highlightcolor}$\result{0.9933}{0.0000}$ \\

& Doorping  & $\result{0.6211}{0.0005}$ & $\result{0.9876}{0.0050}$ & $\result{0.6539}{0.0017}$ & $\result{0.1648}{0.0070}$ & $\result{0.5314}{0.0006}$ & $\result{1.0000}{0.0000}$ \\
& Naive     & $\result{0.6320}{0.0008}$ & $\result{0.1128}{0.0123}$ & $\result{0.6520}{0.0014}$ & $\result{0.1032}{0.0063}$ & $\result{0.5823}{0.0008}$ & $\result{0.0864}{0.0032}$ \\

& Simple    & $\result{0.5837}{0.0003}$ & $\result{0.5896}{0.0118}$ & $\result{0.6519}{0.0010}$ & $\result{0.1424}{0.0081}$ & $\result{0.5373}{0.0009}$ & $\result{0.6736}{0.0317}$ \\
& Relax     & $\result{0.5738}{0.0004}$ & $\result{1.0000}{0.0000}$ & $\result{0.6531}{0.0018}$ & $\result{0.5216}{0.0206}$ & $\result{0.5592}{0.0008}$ & $\result{0.9996}{0.0008}$ \\

\addlinespace

% --- 数据集: SVHN ---
\multirow{6}{*}{\textbf{SVHN}}
& \cellcolor{highlightcolor}InkDrop   & \cellcolor{highlightcolor}$\result{0.7752}{0.0003}$ & \cellcolor{highlightcolor}$\result{0.9976}{0.0000}$ & \cellcolor{highlightcolor}$\result{0.8187}{0.0010}$ & \cellcolor{highlightcolor}$\result{1.0000}{0.0000}$ & \cellcolor{highlightcolor}$\result{0.7570}{0.0003}$ & \cellcolor{highlightcolor}$\result{1.0000}{0.0000}$ \\

& Doorping     & $\result{0.7797}{0.0006}$ & $\result{0.9996}{0.0008}$ & $\result{0.8392}{0.0012}$ & $\result{0.0608}{0.0055}$ & $\result{0.7209}{0.0004}$ & $\result{1.0000}{0.0000}$ \\
& Naive    & $\result{0.7985}{0.0002}$ & $\result{0.1108}{0.0060}$ & $\result{0.8401}{0.0004}$ & $\result{0.1220}{0.0101}$ & $\result{0.7703}{0.0003}$ & $\result{0.1116}{0.0064}$ \\
& Simple     & $\result{0.7479}{0.0002}$ & $\result{0.1104}{0.0067}$ & $\result{0.8415}{0.0009}$ & $\result{0.1140}{0.0081}$ & $\result{0.7589}{0.0005}$ & $\result{0.1140}{0.0051}$ \\
& Relax     & $\result{0.7473}{0.0004}$ & $\result{1.0000}{0.0000}$ & $\result{0.8336}{0.0024}$ & $\result{0.9920}{0.0028}$ & $\result{0.7450}{0.0008}$ & $\result{1.0000}{0.0000}$ \\

\addlinespace
% --- 数据集: FMNIST ---
\multirow{6}{*}{\textbf{FMNIST}}
& \cellcolor{highlightcolor}InkDrop   & \cellcolor{highlightcolor}$\result{0.8883}{0.0005}$ & \cellcolor{highlightcolor}$\result{0.9972}{0.0000}$ & \cellcolor{highlightcolor}$\result{0.8417}{0.0007}$ & \cellcolor{highlightcolor}$\result{0.9856}{0.0011}$ & \cellcolor{highlightcolor}$\result{0.8841}{0.0006}$ & \cellcolor{highlightcolor}$\result{0.9889}{0.0000}$ \\

% & Sneakdoor  & $\result{0.8758}{0.0005}$ & $\result{0.9980}{0.0000}$ & $\result{0.8767}{0.0006}$ & $\result{1.0000}{0.0000}$ & $\result{0.8771}{0.0002}$ & $\result{0.9962}{0.0004}$ \\
& Doorping     & $\result{0.8764}{0.0004}$ & $\result{0.0928}{0.0064}$ & $\result{0.8841}{0.0003}$ & $\result{0.9976}{0.0015}$ & $\result{0.8126}{0.0007}$ & $\result{1.0000}{0.0000}$ \\

& Naive    & $\result{0.8868}{0.0005}$ & $\result{0.0900}{0.0075}$ & $\result{0.8867}{0.0004}$ & $\result{0.0932}{0.0071}$ & $\result{0.8811}{0.0004}$ & $\result{0.0980}{0.0046}$ \\
& Simple     & $\result{0.8682}{0.0004}$ & $\result{0.1784}{0.0053}$ & $\result{0.8786}{0.0002}$ & $\result{0.1588}{0.0069}$ & $\result{0.8801}{0.0002}$ & $\result{0.1508}{0.0119}$ \\

& Relax     & $\result{0.8279}{0.0003}$ & $\result{1.0000}{0.0000}$ & $\result{0.8750}{0.0005}$ & $\result{1.0000}{0.0000}$ & $\result{0.8737}{0.0003}$ & $\result{1.0000}{0.0000}$ \\
\bottomrule
\end{tabular}
\vspace{-9pt}
\end{table*}

\begin{table*}[ht]
\centering
\caption{Attack performance with AlexNetBN as the condensation backbone.}
\label{tab:alexnetbn_final}
\begin{tabular}{clrrrrrr}
\toprule
\multicolumn{2}{>{\columncolor{gray!20}}c}{} &
  \multicolumn{2}{>{\columncolor{dmcolor}}c}{\textbf{DM}} &
  \multicolumn{2}{>{\columncolor{idmcolor}}c}{\textbf{IDM}} &
  \multicolumn{2}{>{\columncolor{damcolor}}c}{\textbf{DAM}} \\
\multirow{-2}{*}{\cellcolor{gray!20}\textbf{Dataset}} &
  \multirow{-2}{*}{\cellcolor{gray!20}\textbf{Method}} &
  \multicolumn{1}{>{\columncolor{dmcolor}}c}{\textbf{CTA}} &
  \multicolumn{1}{>{\columncolor{dmcolor}}c}{\textbf{ASR}} &
  \multicolumn{1}{>{\columncolor{idmcolor}}c}{\textbf{CTA}} &
  \multicolumn{1}{>{\columncolor{idmcolor}}c}{\textbf{ASR}} &
  \multicolumn{1}{>{\columncolor{damcolor}}c}{\textbf{CTA}} &
  \multicolumn{1}{>{\columncolor{damcolor}}c}{\textbf{ASR}} \\
\midrule

\multirow{6}{*}{\textbf{STL10}}
 & \cellcolor{highlightcolor}InkDrop   & \cellcolor{highlightcolor}$\result{0.5553}{0.0040}$ & \cellcolor{highlightcolor}$\result{0.9667}{0.0000}$ & \cellcolor{highlightcolor}$\result{0.6774}{0.0012}$ & \cellcolor{highlightcolor}$\result{0.9067}{0.0133}$ & \cellcolor{highlightcolor}$\result{0.5597}{0.0067}$ & \cellcolor{highlightcolor}$\result{1.0000}{0.0000}$ \\
 % & Sneakdoor  & $\result{0.5617}{0.0013}$ & $\result{0.9925}{0.0000}$ & $\result{0.7235}{0.0015}$ & $\result{0.9860}{0.0017}$ & $\result{0.5839}{0.0009}$ & $\result{0.9623}{0.0031}$ \\
 & Doorping     & $\result{0.5565}{0.0038}$ & $\result{1.0000}{0.0000}$ & $\result{0.6462}{0.0030}$ & $\result{1.0000}{0.0000}$ & $\result{0.5650}{0.0004}$ & $\result{1.0000}{0.0000}$ \\
 & Naive    & $\result{0.5732}{0.0041}$ & $\result{0.1044}{0.0097}$ & $\result{0.7292}{0.0033}$ & $\result{0.1000}{0.0067}$ & $\result{0.6029}{0.0035}$ & $\result{0.1008}{0.0097}$ \\
 & Simple     & $\result{0.5436}{0.0022}$ & $\result{0.0924}{0.0072}$ & $\result{0.7238}{0.0026}$ & $\result{0.1016}{0.0128}$ & $\result{0.5683}{0.0025}$ & $\result{0.0984}{0.0096}$ \\
 & Relax     & $\result{0.5496}{0.0027}$ & $\result{0.7064}{0.0097}$ & $\result{0.7191}{0.0020}$ & $\result{0.6676}{0.0287}$ & $\result{0.5658}{0.0050}$ & $\result{0.8724}{0.0223}$ \\
\addlinespace

\multirow{6}{*}{\shortstack{\textbf{Tiny}\\\textbf{ImageNet}}}
 & \cellcolor{highlightcolor}InkDrop   & \cellcolor{highlightcolor}$\result{0.4474}{0.0015}$ & \cellcolor{highlightcolor}$\result{1.0000}{0.0000}$ & \cellcolor{highlightcolor}$\result{0.2652}{0.0044}$ & \cellcolor{highlightcolor}$\result{1.0000}{0.0000}$ & \cellcolor{highlightcolor}$\result{0.4048}{0.0062}$ & \cellcolor{highlightcolor}$\result{1.0000}{0.0000}$ \\
 % & Sneakdoor  & $\result{0.4634}{0.0017}$ & $\result{0.9200}{0.0126}$ & $\result{0.2598}{0.0046}$ & $\result{0.8600}{0.0126}$ & $\result{0.4418}{0.0060}$ & $\result{0.9720}{0.0098}$ \\
 & Dooring     & $\result{0.4854}{0.0017}$ & $\result{1.0000}{0.0000}$ & $\result{0.2930}{0.0055}$ & $\result{1.0000}{0.0000}$ & $\result{0.4186}{0.0101}$ & $\result{1.0000}{0.0000}$ \\
 & Naive    & $\result{0.4628}{0.0064}$ & $\result{0.0082}{0.0013}$ & $\result{0.2836}{0.0073}$ & $\result{0.0000}{0.0000}$ & $\result{0.4302}{0.0129}$ & $\result{0.0100}{0.0013}$ \\
 & Simple     & $\result{0.4566}{0.0034}$ & $\result{0.0110}{0.0018}$ & $\result{0.3372}{0.0063}$ & $\result{0.0534}{0.0082}$ & $\result{0.4434}{0.0069}$ & $\result{0.0128}{0.0023}$ \\
 & Relax     & $\result{0.4486}{0.0029}$ & $\result{0.8354}{0.0167}$ & $\result{0.3126}{0.0074}$ & $\result{0.7594}{0.0577}$ & $\result{0.4410}{0.0038}$ & $\result{0.7870}{0.0273}$ \\
 \addlinespace

\multirow{6}{*}{\textbf{CIFAR10}}
 & \cellcolor{highlightcolor}InkDrop   & \cellcolor{highlightcolor}$\result{0.6037}{0.0042}$ & \cellcolor{highlightcolor}$\result{0.9773}{0.0053}$ & \cellcolor{highlightcolor}$\result{0.7263}{0.0016}$ & \cellcolor{highlightcolor}$\result{1.0000}{0.0000}$ & \cellcolor{highlightcolor}$\result{0.5941}{0.0023}$ & \cellcolor{highlightcolor}$\result{0.9967}{0.0000}$ \\
 % & Sneakdoor  & $\result{0.5951}{0.0012}$ & $\result{0.9466}{0.0036}$ & $\result{0.7003}{0.0024}$ & $\result{0.9462}{0.0025}$ & $\result{0.6061}{0.0007}$ & $\result{0.7214}{0.0128}$ \\
 & Doorping     & $\result{0.5046}{0.0009}$ & $\result{1.0000}{0.0000}$ & $\result{0.6387}{0.0026}$ & $\result{1.0000}{0.0000}$ & $\result{0.5646}{0.0003}$ & $\result{1.0000}{0.0000}$ \\
 & Naive    & $\result{0.6077}{0.0019}$ & $\result{0.0928}{0.0111}$ & $\result{0.7392}{0.0017}$ & $\result{0.1044}{0.0086}$ & $\result{0.6089}{0.0003}$ & $\result{0.0964}{0.0101}$ \\
 & Simple     & $\result{0.5807}{0.0012}$ & $\result{0.1828}{0.0127}$ & $\result{0.7268}{0.0008}$ & $\result{0.1456}{0.0094}$ & $\result{0.5843}{0.0010}$ & $\result{0.2040}{0.0236}$ \\
 & Relax     & $\result{0.6030}{0.0007}$ & $\result{0.7040}{0.0219}$ & $\result{0.2515}{0.0022}$ & $\result{0.6356}{0.0239}$ & $\result{0.5905}{0.0016}$ & $\result{0.9776}{0.0043}$ \\
\addlinespace

\multirow{6}{*}{\textbf{SVHN}}
 & \cellcolor{highlightcolor}InkDrop   & \cellcolor{highlightcolor}$\result{0.8049}{0.0003}$ & \cellcolor{highlightcolor}$\result{0.9941}{0.0000}$ & \cellcolor{highlightcolor}$\result{0.8846}{0.0006}$ & \cellcolor{highlightcolor}$\result{0.9796}{0.0027}$ & \cellcolor{highlightcolor}$\result{0.8079}{0.0002}$ & \cellcolor{highlightcolor}$\result{0.9863}{0.0000}$ \\
 % & Sneakdoor  & $\result{0.6215}{0.0202}$ & $\result{1.0000}{0.0000}$ & $\result{0.8801}{0.0010}$ & $\result{0.9662}{0.0007}$ & $\result{0.6719}{0.0058}$ & $\result{0.9997}{0.0001}$ \\
 & Doorping     & $\result{0.7741}{0.0013}$ & $\result{1.0000}{0.0000}$ & $\result{0.7811}{0.0016}$ & $\result{1.0000}{0.0000}$ & $\result{0.5931}{0.0033}$ & $\result{1.0000}{0.0000}$ \\
 & Naive    & $\result{0.6973}{0.0068}$ & $\result{0.1236}{0.0059}$ & $\result{0.8858}{0.0006}$ & $\result{0.1164}{0.0099}$ & $\result{0.7007}{0.0021}$ & $\result{0.1124}{0.0077}$ \\
 & Simple     & $\result{0.4844}{0.0097}$ & $\result{0.0712}{0.0045}$ & $\result{0.8797}{0.0008}$ & $\result{0.1176}{0.0077}$ & $\result{0.6932}{0.0062}$ & $\result{0.0924}{0.0069}$ \\
 & Relax     & $\result{0.6723}{0.0092}$ & $\result{0.9776}{0.0074}$ & $\result{0.8736}{0.0008}$ & $\result{0.9996}{0.0008}$ & $\result{0.6921}{0.0028}$ & $\result{0.9956}{0.0027}$ \\
\addlinespace
 
\multirow{6}{*}{\textbf{FMNIST}}
 & \cellcolor{highlightcolor}InkDrop   & \cellcolor{highlightcolor}$\result{0.8088}{0.0035}$ & \cellcolor{highlightcolor}$\result{0.9833}{0.0000}$ & \cellcolor{highlightcolor}$\result{0.8569}{0.0004}$ & \cellcolor{highlightcolor}$\result{0.9844}{0.0014}$ & \cellcolor{highlightcolor}$\result{0.8171}{0.0021}$ & \cellcolor{highlightcolor}$\result{1.0000}{0.0000}$ \\
 % & Sneakdoor  & $\result{0.8218}{0.0001}$ & $\result{1.0000}{0.0000}$ & $\result{0.8437}{0.0010}$ & $\result{0.9784}{0.0022}$ & $\result{0.8307}{0.0026}$ & $\result{1.0000}{0.0000}$ \\
 & Doorping     & $\result{0.6356}{0.0047}$ & $\result{1.0000}{0.0000}$ & $\result{0.7364}{0.0011}$ & $\result{1.0000}{0.0000}$ & $\result{0.7579}{0.0029}$ & $\result{1.0000}{0.0000}$ \\
 & Naive    & $\result{0.8435}{0.0010}$ & $\result{0.0904}{0.0098}$ & $\result{0.8578}{0.0008}$ & $\result{0.1132}{0.0030}$ & $\result{0.8208}{0.0022}$ & $\result{0.1000}{0.0028}$ \\
 & Simple     & $\result{0.8122}{0.0064}$ & $\result{0.9516}{0.0090}$ & $\result{0.8489}{0.0012}$ & $\result{0.2308}{0.0281}$ & $\result{0.8063}{0.0023}$ & $\result{0.4816}{0.1284}$ \\
 & Relax     & $\result{0.8155}{0.0028}$ & $\result{1.0000}{0.0000}$ & $\result{0.8558}{0.0010}$ & $\result{0.7188}{0.0148}$ & $\result{0.8108}{0.0016}$ & $\result{1.0000}{0.0000}$ \\
\bottomrule
\end{tabular}
\vspace{-9pt}
\end{table*}

\subsection{Evaluation Metrics}
We evaluate the attack across three primary dimensions: attack success rate (ASR), clean task accuracy (CTA), and stealthiness (STH). Following prior work~\cite{nips24waveattack}, STH is quantified using three complementary metrics. Firstly, Peak Signal-to-Noise Ratio (PSNR) measures pixel-level similarity between clean and triggered samples, with higher values indicating lower visual distortion. Secondly, Structural Similarity Index Measure (SSIM) assesses structural fidelity, where values closer to 1 reflect stronger perceptual alignment. Thirdly, Inception Score (IS) captures the KL divergence between the predicted class distribution of an individual sample and the marginal distribution, serving as an indicator of semantic recognizability. For interpretability, we define a transformed stealth score as $\text{IS}^{\dagger} = (10^{-3} - \text{IS})\mathrm{e}^{-4}$, such that higher values correspond to greater stealth and lower detectability. % Collectively, these metrics provide a comprehensive assessment of both the visual subtlety and semantic obfuscation introduced by the trigger.

\subsection{Experimental Results}
\subsubsection{Overall Attack Performance}
We begin by evaluating the overall attack performance of each backdoor method in balancing three fundamental objectives: ASR, CTA, and STH. To illustrate the trade-offs among these competing factors, we present radar charts (Figure~\ref{fig:radar}) that depict the normalized scores of each approach across all evaluation dimensions. These visualizations provide a unified view of how well each attack achieves high ASR, preserves CTA, and maintains visual STH. While existing methods exhibit reasonable performance in specific settings, they often fail to consistently inject effective backdoors across diverse datasets and condensation strategies. Their attack success is inconsistent, particularly under varying levels of visual complexity.  In contrast, InkDrop consistently achieves high ASR across all evaluated datasets and condensation strategies, demonstrating superior robustness and adaptability to varying synthetic training distributions.

Moreover, although some existing methods achieve a reasonable balance between ASR and CTA, they often compromise on stealth.  As shown in Figure~\ref{fig:radar}, their performance on perceptual metrics such as PSNR, SSIM, and $\text{IS}^{\dagger}$ is notably weaker, indicating lower visual fidelity and higher detectability of the injected triggers. In comparison, InkDrop achieves a more favorable balance, embedding imperceptible perturbations that preserve benign behavior while sustaining attack efficacy. These findings underscore the necessity of jointly optimizing stealthiness, model utility, and attack effectiveness when designing backdoor attacks within dataset condensation frameworks.

\begin{table*}[ht]
\centering
\small
\caption{Cross-Architecture Attack Transferability with ConvNet Synthesizer.}
\label{tab:cross-arch-results}
\begin{tabular}{clrrrrrr}
\toprule
\multicolumn{2}{>{\columncolor{gray!20}}c}{} &
  \multicolumn{2}{>{\columncolor{dmcolor}}c}{\textbf{DM}} &
  \multicolumn{2}{>{\columncolor{idmcolor}}c}{\textbf{IDM}} &
  \multicolumn{2}{>{\columncolor{damcolor}}c}{\textbf{DAM}} \\
\multirow{-2}{*}{\cellcolor{gray!20}\textbf{Dataset}} &
  \multirow{-2}{*}{\cellcolor{gray!20}\textbf{Network}} &
  \multicolumn{1}{>{\columncolor{dmcolor}}c}{\textbf{CTA}} &
  \multicolumn{1}{>{\columncolor{dmcolor}}c}{\textbf{ASR}} &
  \multicolumn{1}{>{\columncolor{idmcolor}}c}{\textbf{CTA}} &
  \multicolumn{1}{>{\columncolor{idmcolor}}c}{\textbf{ASR}} &
  \multicolumn{1}{>{\columncolor{damcolor}}c}{\textbf{CTA}} &
  \multicolumn{1}{>{\columncolor{damcolor}}c}{\textbf{ASR}} \\
\midrule
\multirow{3}{*}{\textbf{STL10}} & AlexNetBN
  & $\result{0.6014}{0.0015}$ & $\result{1.0000}{0.0000}$
  & $\result{0.6568}{0.0017}$ & $\result{1.0000}{0.0000}$
  & $\result{0.6312}{0.0010}$ & $\result{1.0000}{0.0000}$ \\
 & ResNet18
  & $\result{0.4498}{0.0010}$ & $\result{1.0000}{0.0000}$
  & $\result{0.6246}{0.0014}$ & $\result{1.0000}{0.0000}$
  & $\result{0.4267}{0.0007}$ & $\result{1.0000}{0.0000}$ \\
 & VGG11
  & $\result{0.5772}{0.0010}$ & $\result{1.0000}{0.0000}$
  & $\result{0.6594}{0.0005}$ & $\result{1.0000}{0.0000}$
  & $\result{0.5482}{0.0003}$ & $\result{1.0000}{0.0000}$ \\
\addlinespace
\multirow{3}{*}{\shortstack{\textbf{Tiny}\\\textbf{ImageNet}}} & AlexNetBN
  & $\result{0.5328}{0.0023}$ & $\result{1.0000}{0.0000}$
  & $\result{0.5472}{0.0035}$ & $\result{1.0000}{0.0000}$
  & $\result{0.5594}{0.0043}$ & $\result{1.0000}{0.0000}$ \\
 & ResNet18
  & $\result{0.4508}{0.0016}$ & $\result{1.0000}{0.0000}$
  & $\result{0.4826}{0.0037}$ & $\result{1.0000}{0.0000}$
  & $\result{0.4454}{0.0015}$ & $\result{1.0000}{0.0000}$ \\
 & VGG11
  & $\result{0.4692}{0.0013}$ & $\result{1.0000}{0.0000}$
  & $\result{0.5324}{0.0016}$ & $\result{1.0000}{0.0000}$
  & $\result{0.4456}{0.0021}$ & $\result{1.0000}{0.0000}$ \\
\addlinespace
\multirow{3}{*}{\textbf{CIFAR10}} & AlexNetBN
  & $\result{0.6209}{0.0013}$ & $\result{1.0000}{0.0000}$
  & $\result{0.6922}{0.0008}$ & $\result{1.0000}{0.0000}$
  & $\result{0.6240}{0.0016}$ & $\result{1.0000}{0.0000}$ \\
 & ResNet18
  & $\result{0.5566}{0.0002}$ & $\result{1.0000}{0.0000}$
  & $\result{0.6563}{0.0004}$ & $\result{1.0000}{0.0000}$
  & $\result{0.5244}{0.0003}$ & $\result{1.0000}{0.0000}$ \\
 & VGG11
  & $\result{0.5827}{0.0002}$ & $\result{1.0000}{0.0000}$
  & $\result{0.6538}{0.0002}$ & $\result{1.0000}{0.0000}$
  & $\result{0.5423}{0.0002}$ & $\result{1.0000}{0.0000}$ \\
\addlinespace
\multirow{3}{*}{\textbf{SVHN}} & AlexNetBN
  & $\result{0.7540}{0.0006}$ & $\result{1.0000}{0.0000}$
  & $\result{0.8315}{0.0008}$ & $\result{1.0000}{0.0000}$
  & $\result{0.7820}{0.0009}$ & $\result{1.0000}{0.0000}$ \\
 & ResNet18
  & $\result{0.8074}{0.0002}$ & $\result{1.0000}{0.0000}$
  & $\result{0.8348}{0.0005}$ & $\result{1.0000}{0.0000}$
  & $\result{0.7949}{0.0003}$ & $\result{1.0000}{0.0000}$ \\
 & VGG11
  & $\result{0.8263}{0.0001}$ & $\result{1.0000}{0.0000}$
  & $\result{0.8402}{0.0001}$ & $\result{1.0000}{0.0000}$
  & $\result{0.7970}{0.0001}$ & $\result{1.0000}{0.0000}$ \\
% \addlinespace
% \multirow{3}{*}{\textbf{FMNIST}} & AlexNetBN
%   & $\result{0.8739}{0.0009}$ & $\result{1.0000}{0.0000}$
%   & $\result{0.8354}{0.0004}$ & $\result{1.0000}{0.0000}$
%   & $\result{0.8781}{0.0003}$ & $\result{1.0000}{0.0000}$ \\
%  & ResNet18
%   & $\result{0.8817}{0.0003}$ & $\result{1.0000}{0.0000}$
%   & $\result{0.8398}{0.0002}$ & $\result{1.0000}{0.0000}$
%   & $\result{0.8804}{0.0004}$ & $\result{1.0000}{0.0000}$ \\
%  & VGG11
%   & $\result{0.8802}{0.0003}$ & $\result{1.0000}{0.0000}$
%   & $\result{0.8322}{0.0005}$ & $\result{1.0000}{0.0000}$
%   & $\result{0.8793}{0.0000}$ & $\result{1.0000}{0.0000}$ \\
\bottomrule
\end{tabular}
\vspace{-9pt}
\end{table*}

\begin{table*}[ht]
\centering
\small
\caption{Cross-Architecture Attack Transferability with AlexNetBN Synthesizer.}
\label{tab:cross-arch-alexnetbn}
\begin{tabular}{clrrrrrr}
\toprule
\multicolumn{2}{>{\columncolor{gray!20}}c}{} &
  \multicolumn{2}{>{\columncolor{dmcolor}}c}{\textbf{DM}} &
  \multicolumn{2}{>{\columncolor{idmcolor}}c}{\textbf{IDM}} &
  \multicolumn{2}{>{\columncolor{damcolor}}c}{\textbf{DAM}} \\
\multirow{-2}{*}{\cellcolor{gray!20}\textbf{Dataset}} &
  \multirow{-2}{*}{\cellcolor{gray!20}\textbf{Network}} &
  \multicolumn{1}{>{\columncolor{dmcolor}}c}{\textbf{CTA}} &
  \multicolumn{1}{>{\columncolor{dmcolor}}c}{\textbf{ASR}} &
  \multicolumn{1}{>{\columncolor{idmcolor}}c}{\textbf{CTA}} &
  \multicolumn{1}{>{\columncolor{idmcolor}}c}{\textbf{ASR}} &
  \multicolumn{1}{>{\columncolor{damcolor}}c}{\textbf{CTA}} &
  \multicolumn{1}{>{\columncolor{damcolor}}c}{\textbf{ASR}} \\
\midrule
\multirow{3}{*}{\textbf{STL10}} & ConvNet
  & $\result{0.5392}{0.0002}$ & $\result{1.0000}{0.0000}$
  & $\result{0.6768}{0.0007}$ & $\result{1.0000}{0.0000}$
  & $\result{0.5413}{0.0004}$ & $\result{1.0000}{0.0000}$ \\
 & ResNet18
  & $\result{0.4340}{0.0009}$ & $\result{1.0000}{0.0000}$
  & $\result{0.6333}{0.0004}$ & $\result{1.0000}{0.0000}$
  & $\result{0.4808}{0.0011}$ & $\result{1.0000}{0.0000}$ \\
 & VGG11
  & $\result{0.5332}{0.0008}$ & $\result{1.0000}{0.0000}$
  & $\result{0.6596}{0.0004}$ & $\result{1.0000}{0.0000}$
  & $\result{0.5425}{0.0007}$ & $\result{1.0000}{0.0000}$ \\
\addlinespace
\multirow{3}{*}{\shortstack{\textbf{Tiny}\\\textbf{ImageNet}}} & ConvNet
  & $\result{0.4284}{0.0024}$ & $\result{1.0000}{0.0000}$
  & $\result{0.3728}{0.0071}$ & $\result{1.0000}{0.0000}$
  & $\result{0.4492}{0.0020}$ & $\result{0.9667}{0.0000}$ \\
 & ResNet18
  & $\result{0.3980}{0.0030}$ & $\result{1.0000}{0.0000}$
  & $\result{0.3086}{0.0100}$ & $\result{1.0000}{0.0000}$
  & $\result{0.4058}{0.0013}$ & $\result{1.0000}{0.0000}$ \\
 & VGG11
  & $\result{0.4042}{0.0018}$ & $\result{1.0000}{0.0000}$
  & $\result{0.4638}{0.0037}$ & $\result{1.0000}{0.0000}$
  & $\result{0.4122}{0.0016}$ & $\result{0.9667}{0.0000}$ \\
\addlinespace
\multirow{3}{*}{\textbf{CIFAR10}} & ConvNet
  & $\result{0.5836}{0.0008}$ & $\result{1.0000}{0.0000}$
  & $\result{0.6656}{0.0008}$ & $\result{1.0000}{0.0000}$
  & $\result{0.5744}{0.0008}$ & $\result{1.0000}{0.0000}$ \\
 & ResNet18
  & $\result{0.5305}{0.0005}$ & $\result{0.9967}{0.0000}$
  & $\result{0.7011}{0.0002}$ & $\result{1.0000}{0.0000}$
  & $\result{0.5151}{0.0003}$ & $\result{1.0000}{0.0000}$ \\
 & VGG11
  & $\result{0.5479}{0.0000}$ & $\result{1.0000}{0.0000}$
  & $\result{0.6862}{0.0006}$ & $\result{1.0000}{0.0000}$
  & $\result{0.5449}{0.0001}$ & $\result{1.0000}{0.0000}$ \\
\addlinespace
\multirow{3}{*}{\textbf{SVHN}} & ConvNet
  & $\result{0.7951}{0.0004}$ & $\result{1.0000}{0.0000}$
  & $\result{0.8412}{0.0015}$ & $\result{1.0000}{0.0000}$
  & $\result{0.7963}{0.0002}$ & $\result{1.0000}{0.0000}$ \\
 & ResNet18
  & $\result{0.8193}{0.0002}$ & $\result{0.9980}{0.0000}$
  & $\result{0.8964}{0.0007}$ & $\result{1.0000}{0.0000}$
  & $\result{0.8177}{0.0003}$ & $\result{1.0000}{0.0000}$ \\
 & VGG11
  & $\result{0.8253}{0.0001}$ & $\result{1.0000}{0.0000}$
  & $\result{0.8811}{0.0003}$ & $\result{1.0000}{0.0000}$
  & $\result{0.8347}{0.0001}$ & $\result{1.0000}{0.0000}$ \\
% \addlinespace
% \multirow{3}{*}{\textbf{FMNIST}} & ConvNet
%   & $\result{0.8681}{0.0002}$ & $\result{1.0000}{0.0000}$
%   & $\result{0.8292}{0.0014}$ & $\result{1.0000}{0.0000}$
%   & $\result{0.8420}{0.0004}$ & $\result{1.0000}{0.0000}$ \\
%  & ResNet18
%   & $\result{0.8713}{0.0002}$ & $\result{1.0000}{0.0000}$
%   & $\result{0.8413}{0.0003}$ & $\result{1.0000}{0.0000}$
%   & $\result{0.8540}{0.0001}$ & $\result{1.0000}{0.0000}$ \\
%  & VGG11
%   & $\result{0.8743}{0.0001}$ & $\result{1.0000}{0.0000}$
%   & $\result{0.8470}{0.0002}$ & $\result{1.0000}{0.0000}$
%   & $\result{0.8518}{0.0001}$ & $\result{1.0000}{0.0000}$ \\
\bottomrule
\end{tabular}
\vspace{-9pt}
\end{table*}

\begin{figure}
    \centering

    \begin{minipage}[b]{1\linewidth}
        \centering
        \includegraphics[width=\linewidth]{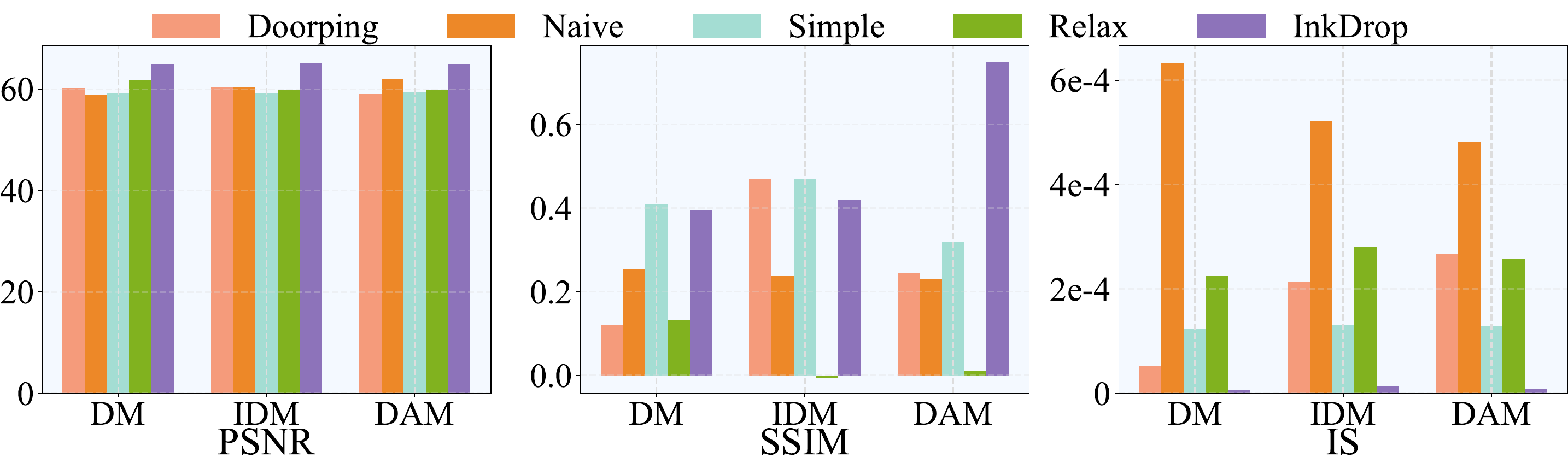}
        (a) STL-10
    \label{subfig:stl-10}
    \end{minipage}
    \hfill
    \begin{minipage}[b]{1\linewidth}
        \centering
        \includegraphics[width=\linewidth]{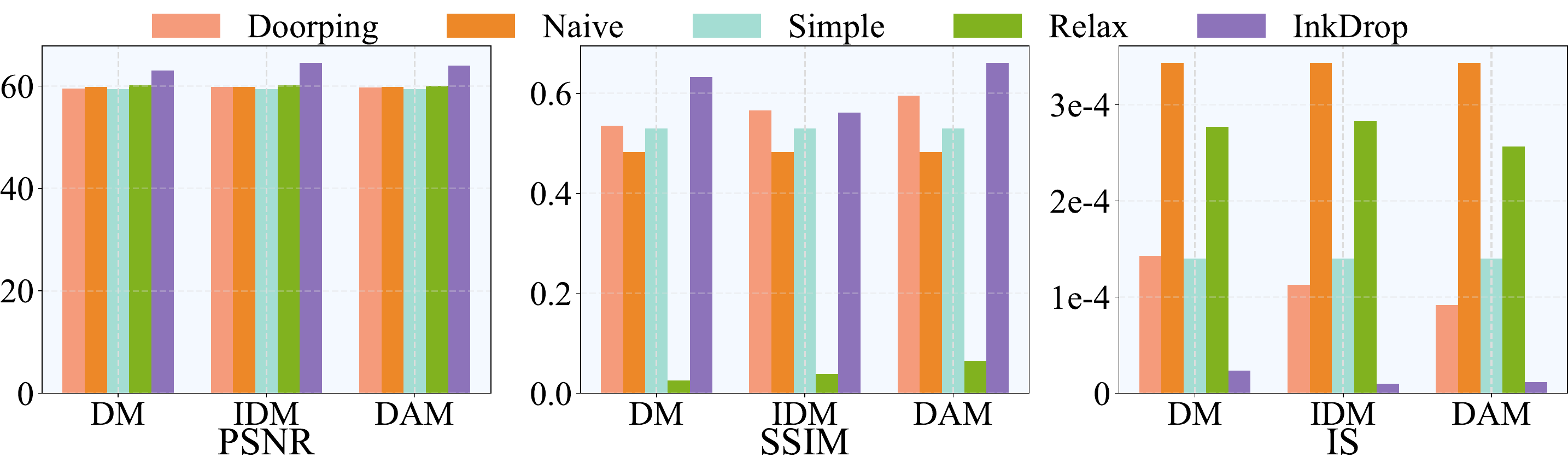}
        (b) Tiny-ImageNet
    \label{subfig:tiny}
    \end{minipage}

    \begin{minipage}[b]{1\linewidth}
        \centering
        \includegraphics[width=\linewidth]{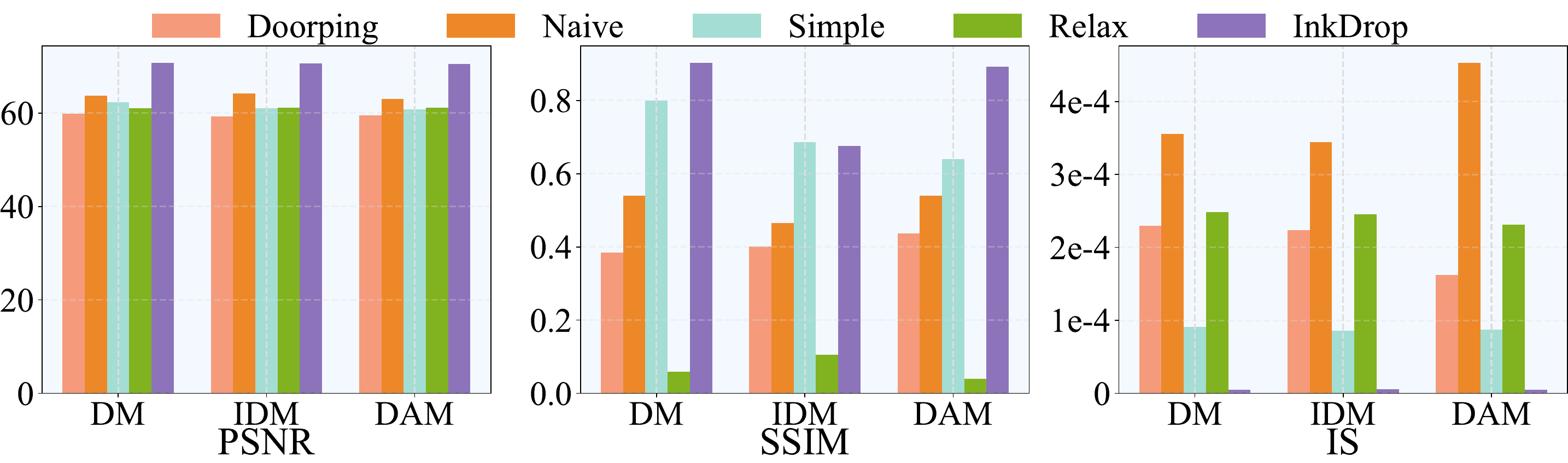}
        (c) CIFAR-10
    \label{subfig:cifar-10}
    \end{minipage}
    % \hfill
    % \begin{minipage}[b]{1\linewidth}
    %     \centering
    %     \includegraphics[width=\linewidth]{journal_svhn_bar_charts.pdf}
    %     (d) SVHN
    % \label{subfig:svhn}
    % \end{minipage}

    \caption{
    Comparison of stealthiness across different datasets using PSNR, SSIM, and IS$^\dagger$ metrics.
    }
    \label{fig:stealthiness}
\vspace{-9pt}
\end{figure}

\subsubsection{Attack Effectiveness on Different Datasets and Synthetic Backbone}
Based on the results shown in Table~\ref{tab:convnet_final} and \ref{tab:alexnetbn_final},  we provide a comprehensive evaluation of attack performance across five datasets, three condensation strategies, and five attack methods. The evaluation focuses on two key metrics: CTA and ASR, which jointly characterize the trade-off between maintaining model utility and achieving effective attack injection. 

Overall, InkDrop consistently achieves near-perfect ASR while maintaining competitive CTA, demonstrating its ability to reliably embed backdoors without compromising model utility across diverse scenarios. Among the baselines, Doorping is the strongest competitor, frequently attaining high ASR and reasonable CTA across datasets. However, it is often accompanied by reduced stealthiness. In contrast, Naive, Simple, and Relax exhibit limited effectiveness, with attack success confined to specific configurations. These methods often fail to achieve a favorable ASR-CTA balance, particularly on more complex datasets or under complex condensation strategies. For instance, under the DM setting on STL10, Naive, Simple, and Relax achieve ASRs of only 0.1488, 0.1032, and 0.0964, respectively, indicating an inability to inject reliable triggers. Moreover, their performance further deteriorates under more advanced strategies such as IDM and DAM, where maintaining both ASR and CTA becomes increasingly difficult. 

The results underscore the consistent superiority of InkDrop, which delivers stable, high-performance backdoor attacks across all datasets and condensation methods. While Doorping remains a viable baseline, its reduced stealthiness limits its practicality in sensitive settings. The remaining methods prove fragile and unreliable in realistic or challenging environments. These findings highlight the necessity of designing backdoor strategies that are not only effective but also robust to variations in data distribution.

\subsubsection{Effectiveness on Cross Architectures}
To evaluate the cross-architecture robustness of InkDrop, we perform experiments in which the condensation and downstream models differ in architectural capacity. This setting mirrors real-world deployment scenarios, where synthetic datasets are often reused across heterogeneous model families. Following~\cite{liu2023backdoor}, we consider four representative architectures: ConvNet, AlexNetBN, VGG11, and ResNet18, and construct all non-identical model pairs by using one network for condensation and a different one for downstream evaluation. As reported in Table~\ref{tab:cross-arch-results} and \ref{tab:cross-arch-alexnetbn}, \textsc{InkDrop} consistently achieves high ASR and maintains reasonable CTA across the majority of model combinations. This indicates that the backdoor signal is not narrowly tailored to the condensation model but instead transfers reliably to diverse architectures. Even in more challenging settings, such as transferring from a lightweight encoder like ConvNet to a deeper network such as ResNet18, InkDrop preserves strong attack efficacy. While slight variations in performance are observed under certain architecture pairs, these fluctuations are minor and do not compromise the overall integrity of the attack. These results underscore the adaptability of InkDrop, demonstrating its ability to maintain both attack strength and model utility under cross-architecture settings.

% 表头分区颜色
\definecolor{tc1color}{rgb}{0.85, 0.95, 0.85}       % 为 DM 定义淡绿色
\definecolor{tc2color}{rgb}{1.0, 0.95, 0.85}      % 为 IDM 定义淡橙色
\definecolor{tc3color}{rgb}{0.9, 0.9, 1.0}        % 为 DAM 定义淡紫色

\begin{table*}[ht]
\centering
\small
\caption{Attack Effectiveness with Varying Target.}
\label{tab:convnet_diff_targets}
\begin{tabular}{clrrrrrr}
\toprule
\multicolumn{2}{>{\columncolor{gray!20}}c}{} &
  \multicolumn{2}{>{\columncolor{tc1color}}c}{\textbf{Target Class 1}} &
  \multicolumn{2}{>{\columncolor{tc2color}}c}{\textbf{Target Class 2}} &
  \multicolumn{2}{>{\columncolor{tc3color}}c}{\textbf{Target Class 3}} \\
\multirow{-2}{*}{\cellcolor{gray!20}\textbf{Dataset}} &
  \multirow{-2}{*}{\cellcolor{gray!20}\textbf{Method}} &
  \multicolumn{1}{>{\columncolor{tc1color}}c}{\textbf{CTA}} &
  \multicolumn{1}{>{\columncolor{tc1color}}c}{\textbf{ASR}} &
  \multicolumn{1}{>{\columncolor{tc2color}}c}{\textbf{CTA}} &
  \multicolumn{1}{>{\columncolor{tc2color}}c}{\textbf{ASR}} &
  \multicolumn{1}{>{\columncolor{tc3color}}c}{\textbf{CTA}} &
  \multicolumn{1}{>{\columncolor{tc3color}}c}{\textbf{ASR}} \\
\midrule
\multirow{3}{*}{\textbf{STL10}}  & DM  
  & $\result{0.5846}{0.0008}$ & $\result{1.0000}{0.0000}$
  & $\result{0.5762}{0.0006}$ & $\result{0.9667}{0.0000}$
  & $\result{0.5836}{0.0007}$ & $\result{0.9667}{0.0000}$ \\
 & IDM 
  & $\result{0.6491}{0.0010}$ & $\result{0.9133}{0.0163}$
  & $\result{0.6454}{0.0011}$ & $\result{1.0000}{0.0000}$
  & $\result{0.6619}{0.0009}$ & $\result{0.9333}{0.0000}$ \\
 & DAM 
  & $\result{0.5355}{0.0008}$ & $\result{1.0000}{0.0000}$
  & $\result{0.5440}{0.0011}$ & $\result{1.0000}{0.0000}$
  & $\result{0.5280}{0.0010}$ & $\result{1.0000}{0.0000}$ \\
\addlinespace
\multirow{3}{*}{\shortstack{\textbf{Tiny}\\\textbf{ImageNet}}} & DM  
  & $\result{0.4910}{0.0017}$ & $\result{1.0000}{0.0000}$
  & $\result{0.4992}{0.0022}$ & $\result{1.0000}{0.0000}$
  & $\result{0.5196}{0.0017}$ & $\result{1.0000}{0.0000}$ \\
 & IDM 
  & $\result{0.5084}{0.0040}$ & $\result{1.0000}{0.0000}$
  & $\result{0.4968}{0.0050}$ & $\result{0.9667}{0.0000}$
  & $\result{0.5140}{0.0021}$ & $\result{0.9667}{0.0000}$ \\
 & DAM 
  & $\result{0.4616}{0.0022}$ & $\result{1.0000}{0.0000}$
  & $\result{0.4710}{0.0044}$ & $\result{1.0000}{0.0000}$
  & $\result{0.4814}{0.0014}$ & $\result{1.0000}{0.0000}$ \\
\addlinespace
\multirow{3}{*}{\textbf{CIFAR10}} & DM  
  & $\result{0.6038}{0.0005}$ & $\result{1.0000}{0.0000}$
  & $\result{0.6037}{0.0002}$ & $\result{1.0000}{0.0000}$
  & $\result{0.6006}{0.0007}$ & $\result{1.0000}{0.0000}$ \\
 & IDM 
  & $\result{0.6494}{0.0016}$ & $\result{1.0000}{0.0000}$
  & $\result{0.6487}{0.0015}$ & $\result{1.0000}{0.0000}$
  & $\result{0.6525}{0.0010}$ & $\result{0.9967}{0.0000}$ \\
 & DAM 
  & $\result{0.5774}{0.0004}$ & $\result{1.0000}{0.0000}$
  & $\result{0.5857}{0.0004}$ & $\result{1.0000}{0.0000}$
  & $\result{0.5817}{0.0008}$ & $\result{1.0000}{0.0000}$ \\
\addlinespace
\multirow{3}{*}{\textbf{SVHN}}   & DM  
  & $\result{0.7692}{0.0005}$ & $\result{1.0000}{0.0000}$
  & $\result{0.7796}{0.0005}$ & $\result{1.0000}{0.0000}$
  & $\result{0.8077}{0.0005}$ & $\result{1.0000}{0.0000}$ \\
 & IDM 
  & $\result{0.8176}{0.0016}$ & $\result{1.0000}{0.0000}$
  & $\result{0.8238}{0.0012}$ & $\result{1.0000}{0.0000}$
  & $\result{0.8241}{0.0013}$ & $\result{1.0000}{0.0000}$ \\
 & DAM 
  & $\result{0.7740}{0.0006}$ & $\result{1.0000}{0.0000}$
  & $\result{0.7595}{0.0006}$ & $\result{1.0000}{0.0000}$
  & $\result{0.7662}{0.0005}$ & $\result{1.0000}{0.0000}$ \\
% \addlinespace
% \multirow{3}{*}{\textbf{FMNIST}} & DM  
%   & $\result{0.8477}{0.0003}$ & $\result{1.0000}{0.0000}$
%   & $\result{0.8385}{0.0003}$ & $\result{1.0000}{0.0000}$
%   & $\result{0.8539}{0.0010}$ & $\result{1.0000}{0.0000}$ \\
%  & IDM 
%   & $\result{0.8334}{0.0012}$ & $\result{1.0000}{0.0000}$
%   & $\result{0.8212}{0.0009}$ & $\result{1.0000}{0.0000}$
%   & $\result{0.8329}{0.0006}$ & $\result{1.0000}{0.0000}$ \\
%  & DAM 
%   & $\result{0.8510}{0.0007}$ & $\result{1.0000}{0.0000}$
%   & $\result{0.8761}{0.0004}$ & $\result{1.0000}{0.0000}$
%   & $\result{0.8582}{0.0004}$ & $\result{1.0000}{0.0000}$ \\
\bottomrule
\end{tabular}
\vspace{-9pt}
\end{table*}

\begin{table*}[ht]
\centering
\small
\caption{Individual Contributions of Each Loss Component in InkDrop.}
\label{tab:stl10_ablation}
\begin{tabular*}{0.938\textwidth}{@{\extracolsep{\fill}}lccccc}
\toprule
\multicolumn{1}{c}{\textbf{Loss Func}} &
  \textbf{CTA} & \textbf{ASR} & \textbf{PSNR} & \textbf{SSIM} & \textbf{IS} \\
\midrule
$\mathcal{L}_{contrast}$
  & $\result{0.6048}{0.0003}$ & $\result{0.8667}{0.0000}$
  & $65.7008$ & $0.1559$ & $8.7760\times10^{-6}$ \\
\addlinespace
$\mathcal{L}_{contrast}+\mathcal{L}_{soft}$
  & $\result{0.6071}{0.0009}$ & $\result{1.0000}{0.0000}$
  & $58.9198$ & $-0.0257$ & $5.9028\times10^{-5}$ \\
\addlinespace
$\mathcal{L}_{contrast}+\mathcal{L}_{soft}+\mathcal{L}_{L2}$
  & $\result{0.6081}{0.0002}$ & $\result{1.0000}{0.0000}$
  & $63.5818$ & $0.1028$ & $1.2014\times10^{-5}$ \\
\addlinespace
$\mathcal{L}_{contrast}+\mathcal{L}_{soft}+\mathcal{L}_{L2}+\mathcal{L}_{LPIPS}$
  & $\result{0.6092}{0.0003}$ & $\result{1.0000}{0.0000}$
  & $64.9668$ & $0.1324$ & $6.2835\times10^{-6}$ \\
\bottomrule
\end{tabular*}
\end{table*}

\begin{table*}[ht]
\centering
\small
\caption{Impact of Varying IPC (Image Per Class, Number of Synthetic Samples per Class) on ASR.}
\label{tab:ipc_effect}
\begin{tabular}{lcccccc}
% \begin{tabular}{c r r r r r r}
\toprule
\multicolumn{1}{>{\columncolor{gray!20}}c}{} &
  \multicolumn{2}{>{\columncolor{dmcolor}}c}{\textbf{IPC = 10}} &
  \multicolumn{2}{>{\columncolor{idmcolor}}c}{\textbf{IPC = 20}} &
  \multicolumn{2}{>{\columncolor{damcolor}}c}{\textbf{IPC = 50}} \\
\multirow{-2}{*}{\cellcolor{gray!20}\textbf{Method}} &
  % \multirow{-2}{*}{\cellcolor{gray!20}\ } &
  \multicolumn{1}{>{\columncolor{dmcolor}}c}{\textbf{CTA}} &
  \multicolumn{1}{>{\columncolor{dmcolor}}c}{\textbf{ASR}} &
  \multicolumn{1}{>{\columncolor{idmcolor}}c}{\textbf{CTA}} &
  \multicolumn{1}{>{\columncolor{idmcolor}}c}{\textbf{ASR}} &
  \multicolumn{1}{>{\columncolor{damcolor}}c}{\textbf{CTA}} &
  \multicolumn{1}{>{\columncolor{damcolor}}c}{\textbf{ASR}} \\
\midrule
\textbf{DM} 
  & $\result{0.4469}{0.0006}$ & $\result{1.0000}{0.0000}$
  & $\result{0.5072}{0.0005}$ & $\result{0.9667}{0.0000}$
  & $\result{0.5805}{0.0008}$ & $\result{1.0000}{0.0000}$ \\
\addlinespace
\textbf{IDM} 
  & $\result{0.5851}{0.0010}$ & $\result{0.9333}{0.0000}$
  & $\result{0.6223}{0.0015}$ & $\result{0.9667}{0.0000}$
  & $\result{0.6476}{0.0007}$ & $\result{0.9333}{0.0000}$ \\
\addlinespace
\textbf{DAM} 
  & $\result{0.3909}{0.0003}$ & $\result{0.9667}{0.0000}$
  & $\result{0.4508}{0.0008}$ & $\result{1.0000}{0.0000}$
  & $\result{0.5350}{0.0011}$ & $\result{1.0000}{0.0000}$ \\
\bottomrule
\end{tabular}
\vspace{-9pt}
\end{table*}

\begin{figure}
    \centering
    \includegraphics[width=1\linewidth]{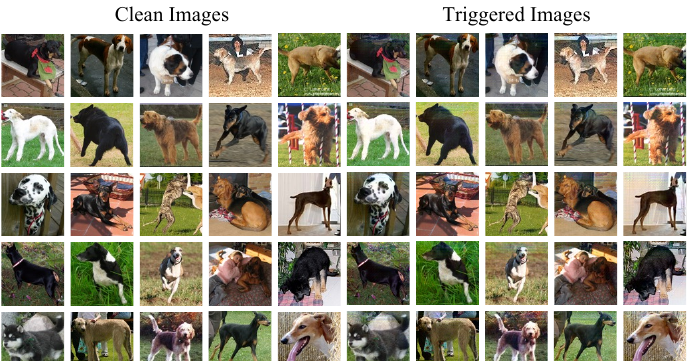}
    \caption{Visual comparison between clean and triggered samples on STL10.}
    \label{fig:visual-comparison}
    \vspace{-9pt}
\end{figure}

\subsubsection{Evaluation of Stealthiness}
As shown in Figure~\ref{fig:stealthiness}, InkDrop exhibits superior stealthiness performance across all condensation strategies, consistently surpassing baseline methods in perceptual fidelity. The generated perturbations preserve high visual quality, as reflected in results across all three stealthiness metrics. This indicates that the injected triggers are subtle and non-disruptive, maintaining both local structure and global perceptual coherence. Unlike prior methods, which often introduce conspicuous artifacts or semantic inconsistencies, InkDrop produces natural-looking modifications that closely resemble clean samples. While certain baselines achieve moderate scores on individual stealth indicators, they compromise on ASR. In contrast, InkDrop achieves a favorable balance, embedding imperceptible triggers without compromising ASR or CTA. These findings underscore the effectiveness of jointly optimizing for stealthiness and functionality during attack construction. 

Moreover, Figure~\ref{fig:visual-comparison} provides a side-by-side visualization of clean samples and their corresponding triggered versions synthesized by InkDrop on the STL10 dataset. The injected perturbations are visually subtle, preserving both local texture and global structure with minimal perceptual deviation. Notably, the modifications introduce no apparent artifacts or semantic inconsistencies, highlighting the effectiveness of the stealth-aware objective. This visual evidence supports the claim that InkDrop generates imperceptible triggers while maintaining high visual fidelity.

\subsubsection{Attack Effectiveness with Varying Target}
To further evaluate the robustness and adaptability of InkDrop, we conduct comprehensive experiments in which the target class is randomly selected across various datasets and condensation strategies, as shown in Table~\ref{tab:convnet_diff_targets}. This setting simulates realistic scenarios where the attacker can flexibly choose an effective target class rather than relying on fixed assumptions. Despite the variability introduced by different target classes, InkDrop consistently achieves high ASR. These results demonstrate that InkDrop can effectively synthesize trigger patterns that align with diverse class distributions, without relying on class-specific heuristics or manual adjustments. This resilience to target class variability, combined with its compatibility across condensation frameworks, underscores the practicality of InkDrop for real-world deployment, where attackers may seek to optimize impact under dynamic or uncertain conditions. 

\begin{table*}[ht]
\centering
\caption{Attack Effectiveness of InkDrop when applying defenses such as PDB and RNP.}
\label{tab:pdb_rnp}
\begin{tabular}{c r r r r r r r r r r r r}
\toprule
\multicolumn{1}{>{\columncolor{gray!20}}c}{} &
  \multicolumn{4}{>{\columncolor{dmcolor}}c}{\textbf{DM}} &
  \multicolumn{4}{>{\columncolor{idmcolor}}c}{\textbf{IDM}} &
  \multicolumn{4}{>{\columncolor{damcolor}}c}{\textbf{DAM}} \\
\multicolumn{1}{>{\columncolor{gray!20}}c}{} &
  \multicolumn{2}{>{\columncolor{dmcolor}}c}{\cellcolor{dmcolor}\textbf{PDB}} &
  \multicolumn{2}{>{\columncolor{dmcolor}}c}{\cellcolor{dmcolor}\textbf{RNP}} &
  \multicolumn{2}{>{\columncolor{idmcolor}}c}{\cellcolor{idmcolor}\textbf{PDB}} &
  \multicolumn{2}{>{\columncolor{idmcolor}}c}{\cellcolor{idmcolor}\textbf{RNP}} &
  \multicolumn{2}{>{\columncolor{damcolor}}c}{\cellcolor{damcolor}\textbf{PDB}} &
  \multicolumn{2}{>{\columncolor{damcolor}}c}{\cellcolor{damcolor}\textbf{RNP}} \\
 \multirow{-3}{*}{\cellcolor{gray!20}\textbf{Dataset}} & 
  \cellcolor{dmcolor}\textbf{CTA} & \cellcolor{dmcolor}\textbf{ASR} &
  \cellcolor{dmcolor}\textbf{CTA} & \cellcolor{dmcolor}\textbf{ASR} &
  \cellcolor{idmcolor}\textbf{CTA} & \cellcolor{idmcolor}\textbf{ASR} &
  \cellcolor{idmcolor}\textbf{CTA} & \cellcolor{idmcolor}\textbf{ASR} &
  \cellcolor{damcolor}\textbf{CTA} & \cellcolor{damcolor}\textbf{ASR} &
  \cellcolor{damcolor}\textbf{CTA} & \cellcolor{damcolor}\textbf{ASR} \\
\midrule
\textbf{STL10}         & 0.0940 & 0.1000 & 0.4641 & 1.0000 & 0.1658 & 0.9000 & 0.3015 & 0.6333 & 0.0800 & 0.2333 & 0.5343 & 1.0000 \\
\addlinespace
\textbf{Tiny-ImageNet} & 0.0540 & 0.0333 & 0.2100 & 0.0000 & 0.0950 & 0.2333 & 0.0820 & 0.0000 & 0.0390 & 0.1333 & 0.2200 & 0.8667 \\
\addlinespace
\textbf{CIFAR10}       & 0.0723 & 0.1333 & 0.2900 & 0.9933 & 0.1994 & 0.5267 & 0.2407 & 0.5867 & 0.0647 & 0.1033 & 0.3972 & 1.0000 \\
\addlinespace
\textbf{SVHN}          & 0.0657 & 0.1429 & 0.5192 & 1.0000 & 0.2323 & 0.6538 & 0.6082 & 0.9952 & 0.0547 & 0.0799 & 0.6120 & 1.0000 \\
\bottomrule
\end{tabular}
\vspace{-9pt}
\end{table*}

\subsubsection{Ablation Study}
Table~\ref{tab:stl10_ablation} reports an ablation study on the STL10 dataset, evaluating the individual contribution of each loss component in InkDrop. With only the contrastive loss, the model achieves a moderate ASR of 0.8667, indicating initial effectiveness in encoding backdoor behavior. Introducing soft label alignment loss $\mathcal{L}_{\text{soft}}$ elevates ASR to 1, but at the expense of PSNR, SSIM, and IS, reflecting a reduction in visual fidelity. This indicates that the trigger becomes more effective, but also more conspicuous. Adding the L2 regularization term $\mathcal{L}_{\text{L2}}$ improves stealthiness by suppressing the magnitude of excessive perturbation, yielding better stealthiness metrics while preserving attack efficacy. 
The final addition of perceptual loss $\mathcal{L}_{\text{LPIPS}}$ further enhances visual realism without sacrificing ASR or CTA. Collectively, these results demonstrate the complementary nature of loss components: $\mathcal{L}_{\text{contrast}}$ and $\mathcal{L}_{\text{soft}}$ promote effective backdoor embedding, while $\mathcal{L}_{\text{L2}}$ and $\mathcal{L}_{\text{LPIPS}}$ refine stealth and perturbation quality.

\begin{table}[ht]
\centering
\caption{Attack Effectiveness of InkDrop when applying PIXEL and ABS.}
\label{tab:pixel_abs_reasr}
\begin{tabular}{lcccccc}
\toprule
\multicolumn{1}{>{\columncolor{gray!20}}c}{} &
  \multicolumn{2}{>{\columncolor{dmcolor}}c}{\textbf{DM}} &
  \multicolumn{2}{>{\columncolor{idmcolor}}c}{\textbf{IDM}} &
  \multicolumn{2}{>{\columncolor{damcolor}}c}{\textbf{DAM}} \\
\multirow{-2}{*}{\cellcolor{gray!20}\textbf{Dataset}} &
  \multicolumn{1}{>{\columncolor{dmcolor}}c}{\textbf{PIXEL}} &
  \multicolumn{1}{>{\columncolor{dmcolor}}c}{\textbf{ABS}} &
  \multicolumn{1}{>{\columncolor{idmcolor}}c}{\textbf{PIXEL}} &
  \multicolumn{1}{>{\columncolor{idmcolor}}c}{\textbf{ABS}} &
  \multicolumn{1}{>{\columncolor{damcolor}}c}{\textbf{PIXEL}} &
  \multicolumn{1}{>{\columncolor{damcolor}}c}{\textbf{ABS}} \\
\midrule

\textbf{STL10}         & 1.7860 & 0.20 & 0.9122 & 0.18 & 1.6289 & 0.16 \\
\addlinespace
\textbf{Tiny-ImageNet} & 1.1632 & 0.22 & 1.8707 & 0.10 & 1.6165 & 0.10 \\
\addlinespace
\textbf{CIFAR10}       & 1.4333 & 0.44 & 1.9543 & 0.38 & 1.0792 & 0.51 \\
\addlinespace
\textbf{SVHN}          & 1.7061 & 0.30 & 0.8672 & 0.38 & 0.9636 & 0.39 \\
\bottomrule
\end{tabular}
\vspace{-9pt}
\end{table}

\subsubsection{Influence of the Size of Condensation Data}
Table~\ref{tab:ipc_effect} analyzes how varying the number of images per class (IPC) in the distilled dataset influences the attack performance of InkDrop. Across all condensation strategies and IPC settings of 10, 20, and 50, the attack consistently achieves high ASR. This demonstrates that InkDrop reliably embeds effective backdoors independent of the synthetic dataset size. While CTA increases with larger IPC values, the attack remains effective even under constrained data conditions. These results highlight InkDrop's effectiveness in handling different data volumes.

\subsubsection{Robust to Existing Defenses}
To evaluate the robustness of InkDrop against existing defenses, we systematically assess its performance under a comprehensive set of state-of-the-art strategies, including model-level, input-level, and dataset-level defenses (as shown in Table~\ref{tab:pixel_abs_reasr} and \ref{tab:pdb_rnp}). Across all evaluated settings, InkDrop maintains high ASR and competitive CTA, demonstrating strong resilience. For model-level defenses such as PIXEL~\cite{taog2022better}, InkDrop consistently yields anomaly scores below detection thresholds, indicating that the injected triggers do not induce recognizable deviations. Input-level defenses like ABS~\cite{abs}, which aim to reconstruct clean inputs, exhibit limited effectiveness; the regenerated inputs fail to remove the backdoor effect, resulting in sustained attack success. Dataset-level methods such as RNP~\cite{li2023reconstructive} and PDB~\cite{wei2024mitigating}, which attempt to filter poisoned samples, occasionally suppress ASR but simultaneously reduce CTA, reflecting a substantial trade-off. These results underscore the difficulty of defending against the subtle, input-adaptive perturbations introduced by InkDrop and highlight its robustness across diverse defense paradigms.

\section{Conclusion}
This paper presents InkDrop, a stealthy backdoor attack tailored for dataset condensation. Motivated by the insight that samples near decision boundaries are inherently susceptible to class shifts, InkDrop selectively perturbs these boundary-adjacent instances. It employs a learnable, instance-specific trigger generator trained via a multi-objective loss that integrates contrastive alignment, distribution matching, and perceptual consistency. Comprehensive evaluations across diverse datasets, condensation strategies, and defense mechanisms demonstrate that InkDrop consistently achieves a favorable balance between attack effectiveness, model utility, and stealthiness. These findings highlight the crucial role of decision-boundary sensitivity and stealth-aware design in developing effective and resilient backdoor attacks under realistic constraints.

\bibliographystyle{IEEEtran}
\bibliography{refs}

\vfill

\end{document}